\documentclass[preprint,12pt]{elsarticle}

\usepackage{amssymb}
\usepackage[utf8]{inputenc} 
\usepackage[T1]{fontenc}    
\usepackage[colorlinks=True]{hyperref}       
\usepackage{url}            
\usepackage{booktabs}       
\usepackage{amsfonts}       
\usepackage{nicefrac}       
\usepackage{microtype}      
\usepackage{graphicx}
\usepackage{lineno}
\usepackage{amsmath}
\usepackage{tabulary}
\usepackage{ulem}
\usepackage{tabularx,ragged2e,caption}
\usepackage{breqn}
\usepackage{siunitx}
\usepackage[inline]{enumitem}
\usepackage{xcolor}
\usepackage{mathtools}
\usepackage{mathrsfs}
\usepackage{hyperref}
\usepackage{bbm}
\usepackage{lipsum}
\usepackage{subfig}
\usepackage{booktabs,floatrow}
\usepackage{wrapfig}
\usepackage{adjustbox}
\usepackage{longtable}
\usepackage{multicol}
\usepackage{scalerel,stackengine}
\usepackage{scrextend}

\stackMath
\newcommand\reallywidehat[1]{%
\savestack{\tmpbox}{\stretchto{%
  \scaleto{%
    \scalerel*[\widthof{\ensuremath{#1}}]{\kern-.6pt\bigwedge\kern-.6pt}%
    {\rule[-\textheight/2]{1ex}{\textheight}}
  }{\textheight}%
}{0.5ex}}%
\stackon[1pt]{#1}{\tmpbox}%
}
\DeclareMathOperator{\FC}{\textsc{FC}}

\DeclareMathOperator{\NBEATS}{\textsc{NBEATS}}

\journal{Elsevier Journal}

\begin{document}

\begin{frontmatter}



\title{Neural Forecasting at Scale}

\author[inst1]{Philippe Chatigny}
\author[inst1]{Shengrui Wang}
\author[inst3]{Jean-Marc Patenaude}
\author[inst2]{Boris N. Oreshkin}

\affiliation[inst1]{organization={University of Sherbrooke},
            city={Sherbrooke},
            state={QC},
            country={Canada}
            }
\affiliation[inst2]{organization={Unity Technologies, Labs},
            city={Montreal},
            state={QC},
            country={Canada}
            }
\affiliation[inst3]{organization={Laplace Insights},
            city={Sherbrooke},
            state={QC},
            country={Canada}
            }

\begin{abstract}
We study the problem of efficiently scaling ensemble-based deep neural networks for multi-step time series (TS) forecasting on a large set of time series. Current state-of-the-art deep ensemble models have high memory and computational requirements, hampering their use to forecast millions of TS in practical scenarios. We propose N-BEATS(P), a global parallel variant of the N-BEATS model designed to allow simultaneous training of multiple univariate TS forecasting models. Our model addresses the practical limitations of related models, reducing the training time by half and memory requirement by a factor of 5, while keeping the same level of accuracy in all TS forecasting settings. We have performed multiple experiments detailing the various ways to train our model and have obtained results that demonstrate its capacity to generalize in various forecasting conditions and setups.
\end{abstract}







\end{frontmatter}

\section{Introduction}
\label{sec:Introduction}
In the past few years, abundant evidence has emerged suggesting that deep neural networks (DNN) constitute an effective modeling framework for solving time series (TS) forecasting problems. DNN models have been shown to produce state-of-the-art forecasts when large homogeneous datasets with multiple observations are available~\cite{makridakis2020forecasting}. The success of DNN is largely accounted for by two factors: (i) the cross-learning on multiple time-series\footnote{Cross-learning is the approach were a single model is trained on multiple TS. The model assumes that all TS follow the same process and that each TS are independent samples of this process. A notable instance of such model is the FFORMA model \cite{montero2020fforma}.} \cite{M5Makridakis,makridakis2020m4,semenoglou2021investigating} and (ii) the use of over-specified large capacity ensemble models\footnote{Over-specified large capacity ensembles refers to ensemble of models (DNN here) where each model have high number of parameters, perhaps larger than what would be needed to get a good training error. We refer the reader to recent empirical \cite{choromanska2015loss,nakkiran2019deep} and theoretical \cite{dauphin2014identifying, bach2017breaking} evidences that indicates that larger networks may indeed be easier to train to achieve better results.}. However, the high computational requirements of such models in comparison to statistical models have raised concerns regarding their applicability in practical scenarios~\cite{M5Makridakis}. Indeed, the deployment of a reliable DNN with an automatic training procedure is far more challenging because of this cost and other factors such as optimal architecture and hyperparameter tuning which various authors discused in previous studies \cite{chen2020comprehensive,paleyes2020challenges}. These factors can be summarized by the following. 

The need to render these methods more efficient has been pointed out multiple times \cite{M5Makridakis,slides_on_Forecasting_Amazon} and is one of the core challenges that must be solved to democratize their use. Currently, they require much time, specialized hardware and energy to train and deploy. Besides their model size, which can render their use cumbersome, re-training these models at every forecast for different TS is not viable for most organizations given their training time. Reducing memory requirements and computational cost, as well as offering model that are “ready-to-use” once trained are key aspects to improve upon to make these model more accessible to smaller organizations who neither have the money nor the data to support frequent retraining. 

Only recently has some work been done to evaluate how to generalize these models to multiple types of TS when trained on public datasets that cover various TS settings while maintaining an acceptable level of accuracy within a zero-shot regime \cite{oreshkin2020meta}, i.e. to train a neural network on a source TS dataset and deploy it on a different target TS dataset without retraining, which provides a more efficient and reliable solution to forecast at scale than its predecessor even in difficult forecasting conditions, or in a few-shot learning regime, i.e., by fine-tuning the model to the target dataset of interest \cite{hooshmand2019energy,gupta2020transfer}. In the ensemble case, the cost of runing these models is amplified since most of the current top-performing models rely on independent training of ensemble members. Producing forecasts with a small ensemble size without affecting accuracy is of great interest for smaller organization.

On the other hand, there are plentiful examples of successful deployment of neural networks in large-scale TS forecasting. It appears that the benefits of using such models in practice definitely outweigh the associated costs and difficulties \cite{montero2021principles}. First, such models are \emph{scalable}: a neural TS forecasting model usually performs better as the scale of the data used to train it increases. This has been observed in various TS competitions \cite{M5Makridakis,makridakis2018m4,google_kaggle_comp}. Second, they can be \emph{reusable}: we can reuse a model to produce forecasts over multiple TS \cite{oreshkin2020meta} not observed during training faster and produce forecast in real-time. They also offer \emph{flexibility}: It is typicaly easier to adjust a DNN-based model for handling missing values \cite{rubanova2019latent}, adjusting its parameters based on custom business/scientific objectives \cite{smyl_m4_2018} or considering multi-modal (various source and representation of data such as text, video, etc.) within a single end-to-end model \cite{ekambaram2020attention}.

These benefits made neural TS forecasting models popular and even mainstream in various settings. In fact, the prevalent use of neural networks manifests a paradigm shift in data-driven forecasting techniques, with fully-automated models being the de-facto standard in organizations that that can afford DNN based forecasting workflows. Many examples of DNN for TS exist. Some of the largest online retail platforms are using neural networks to forecast product demand for millions of retail items 
\cite{salinas2020deepar,bose2017probabilistic}. AutoML solutions with heavy use of DNNs like \cite{googlaiblog} are being used in various settings and have been demonstrated to be very competitive \cite{M5Makridakis} with almost no human involvement. Some companies that need to allocate a large pool of resources in different environments are using neural networks to anticipate required resources for different periods of the day \cite{bell2018forecasting,uberpresentation}. Large capital markets companies are using neural networks to predict the future movement of assets \cite{banushev_barclay_2021} via a process that links the trade-generating strategies with notifications and trade automation from these forecasts.\footnote{For interested reader, this special kind of forecasting is known as asset pricing~\cite{cochrane2009asset, chatigny2021asset}. In this setup we are interested in modelizing the relationship between systematic risk factor and expected excess return of assets over the market and ultimately build an optimal portfolio. This goes out of the scope of this paper. We focus on forecasting any TS, regardless if is an asset, solely based on its historical values.}  However, these approaches are often not accessible to smaller organization because of their cost to opperate. We propose to tackle the problems within a single approach to facilitate the use of DNN for TS forecasting at scale.

\section{Related Work}
\label{sec:Related Works}

\textbf{TS forecasting models:} Traditional local, univariate models for TS forecasting include the autoregressive integrated moving average (ARIMA) model \cite{box2015time}, exponential smoothing methods (HOLT, ETS, DAMPED, SES) \cite{hyndman2008forecasting,makridakis2020m4} decomposition-based approaches, including the THETA model~\cite{assimakopoulos2000theta}, and autoregressive (AR) models with time-varying coefficients as in \cite{hamilton1989new,prado2000bayesian}. Global univariate TS models that rely on deep neural networks (DNNs) have been proposed recently as alternatives to these models such as DEEP-STATE \cite{rangapuram2018deep}, DEEP-AR \cite{salinas2020deepar} and more recently \textit{Transformer}-based models \cite{li2019enhancing,wu2020deep}. In contrast to the traditional approaches, they can be trained with multiple independent TS simultaneously and handle non-stationary TS without preprocessing steps. One of the key difference between these two class of model is are how they approach the forecasting problem. Traditonal models typically learn from TS locally, by considering each TS as a separate regression task and fitting a function to each (local model) whereas DNNs do so by fitting a single function to multiple TS (global model) \cite{montero2021principles}.

Some concerns have been raised regarding machine learning (ML) publications claiming satisfactory accuracy without adequate comaprison with the well-established statistical baselines and using inappropriate criteria often leading to misleading results~\cite{makridakis2018statistical}. It is inspiring to see that recent ML publications such as \cite{rangapuram2018deep,salinas2020deepar,oreshkin2019n} have largely solved these problems by following more rigorous evaluation protocols and baseline comparisons.

\textbf{Ensemble methods:} Combining multiple models is often a more straightforward strategy to produce accurate forecasts than finding the best parameterization for one particular model~\cite{clemen1989combining,timmermann2006forecast}. Recently, both M4 and M5 forecasting competitions have empirically confirmed the accuracy of ensemble methods \cite{makridakis2018m4,M5Makridakis}. Notable instances of model for univariate TS forecasting include that use this method FFORMA~\cite{montero2020fforma} (second entry in M4), ES-RNN~\cite{smyl_hybrid_2020} (first entry in M4) and subsequently N-BEATS\footnote{N-BEATS was not part of the M4 competition, and attained state-of-the-art results on M4 benchmark ex post facto. N-BEATS was the core part of the second-entry solution in M5 competition\cite{M5Makridakis}.}~\cite{oreshkin2019n}. Because they use ensembling, these models, especially N-BEATS, have high computational and memory complexities, which require specialized infrastructure to accelerate their training and store the trained models~\cite{makridakis2020m4}. For example, the full N-BEATS models consists of 180 individual models. It takes around 11'755 hours to train on the full M4 dataset using 1 NVIDIA GTX 2080Ti GPU. Furthermore, the total size of the models in ensemble is 160 GB, which, depending on the number of training logs and saved snapshots of the model, can increase to over 450 GB. In comparison, the Theta method takes around 7 min to do the same.

\textbf{N-BEATS:} The overall N-BEAT model \cite{oreshkin2019n} is designed to apply signal decomposition of the original TS similar to the “seasonality-trend-level” approach of \cite{cleveland1990stl} but using a fully connected neural networks organized into a set of \textit{blocks}. Each blocks applies a decomposition of the signal it is given as input and make a forecast from this signal and pass the reminder of the signal to the other block. Beside the parmaterization of each block, one as to specify the number of past observations all blocks must consider which we refer as the \textit{lookback} windows. When trained on large datasets, N-BEATS is trained with a bagging proeceduce \cite{breiman1996bagging} on various lookback windows, losses, and subpopulations to produce an ensemble of models. All of these models are independently trained and inflate the parameter size of the ensemble and thus the time to train the complete model.

Hence, the major issues of using these model at scale come down to parameter size of the model, time to train the model and whether or not we can offset the operating costs of DNN based forecasting workflows for ensemble models. This paper seek to reduce the computational complexity gap between classical and neural TS models by proposing a more memory- and computation -efficient version of the N-BEATS model \cite{oreshkin2019n}. Our approach achieves this by re-formulating the original fully-connected N-BEATS architecture as a single kernel convolution, which allows for training multiple models, each with different lookback windows, in parallel on the same GPU while sharing most of the parameters in the network. This leads to reduced ensemble training time and memory footprint as well as reduced ensemble model size, which positively affects the costs of training, querying and storing the resulting ensemble without compromising its accuracy.

Our contributions can be summarized as follows:
\begin{enumerate}[label={\textbf{[\arabic*]}}]
    \item We introduce N-BEATS(P), a multi-head parallelizable N-BEATS architecture that permits the simultaneous training of multiple global TS models. Our model is twice as fast as N-BEATS, has 5 times fewer parameters than its predecessor, and performs at the same level of accuracy on M4 dataset than N-BEATS and generalize well in other TS forecast condition.
    \item Our model is faster to train and more accurate than the top-scoring models of the M4 competition (ES-RNN~\cite{smyl_hybrid_2020} FFORMA~\cite{montero2020fforma}). 
\end{enumerate}

The remainder of this paper is organized as follows. Section~\ref{sec:Problem statement} describes the univariate TS forecasting problem. Section~\ref{sec:Model} presents our modeling approach. Section~\ref{sec:Experimental analysis} outlines empirical evaluation setup and our results. Finally, Section~\ref{sec:Conclusion} presents our conclusions.

\section{Problem Statement}
\label{sec:Problem statement}

We consider the univariate point forecasting problem in discrete time where we have a training dataset of $N$ TS, $\mathscr{D}_{\mathrm{train}}=\{\textbf{X}^{(i)}_{1:T_{i}} \}_{i=1}^N$ and a test dataset of future values of these TS $\mathscr{D}_{\mathrm{eval}}=\{\textbf{Y}^{(i)}_{T_{i}+1:T_{i}+H} \}_{i=1}^N$. The task is to forecast future values of the series, $\textbf{{Y}}^{(i)}_{T_i+1:H} \in \mathbb{R}^{H}$, given a regularly-sampled sequence of past observations, $\textbf{X}^{(i)}_{1:T_{i}} \in \mathbb{R}^{T^{(i)}}$. We use the bold notation to define vectors, matrix and tensor. To solve the task, we define a forecasting function $f_{\theta} : \mathbb{R}^{l} \rightarrow \mathbb{R}^{H}$, parameterized with a set of learnable parameters $\theta \in \Theta \subset \mathbb{R}^{M}$ where $l\leq T_i$ The parameters of the forecasting function can be learned using an empirical risk minimization framework based on the appropriate samples of forecasting function inputs, $\textbf{Z}_{\textrm{in}} \in \mathbb{R}^{l}$, and outputs, $\textbf{Z}_{\textrm{out}} \in \mathbb{R}^{H}$, taken from the training set:
\begin{align}
    \widehat\theta = \arg\min_{\theta \in \Theta} \sum_{\textbf{Z}_{\textrm{in}}, \textbf{Z}_{\textrm{out}} \in \mathscr{D}_{\mathrm{train}}} \mathcal{L}(\textbf{Z}_{\textrm{out}}, f_{\theta}(\textbf{Z}_{\textrm{in}}))
\end{align}
A few remarks are in order regarding the selection of the model input window size $l$. The optimal choice of $l$ is highly data-dependent. In terms of general guidelines, TS with a swiftly changing generating process \cite{vzliobaite2010learning} will favor small values of $l$, as historical information quickly becomes outdated. TS with long seasonality periods will favor larger $l$, as observing at least one and maybe a few seasonality periods may be beneficial for making a more informed forecast. Obviously, several conflicting factors can be at play here and finding a universally optimal solution for all TS does not seem viable. Therefore, $l$ can be treated as a hyperparameter selected on a TS-specific validation set. A more productive and accurate solution would entail using an ensemble of several models, each trained with its own $l$, as in \cite{oreshkin2019n}. However, this solution tends to inflate the ensemble size, and that is the problem we aim to address in this paper. In general, increasing the diversity of an ensemble \cite{zhou2012ensemble} with different forecasting models usually results in the inflation of the ensemble size and computational costs. Therefore, we focus on providing a solution to more effectively parallelize training of the N-BEATS ensemble, which is obviously applicable to situations other than using multi-$l$ ensembles.

\section{Model}
\label{sec:Model}

\subsection{Model architecture}
\label{sec:Model architecture}

\begin{figure}[htbp]
\centering 
\includegraphics[trim={3cm 0.5cm 4cm 2cm},width=0.85\textwidth]{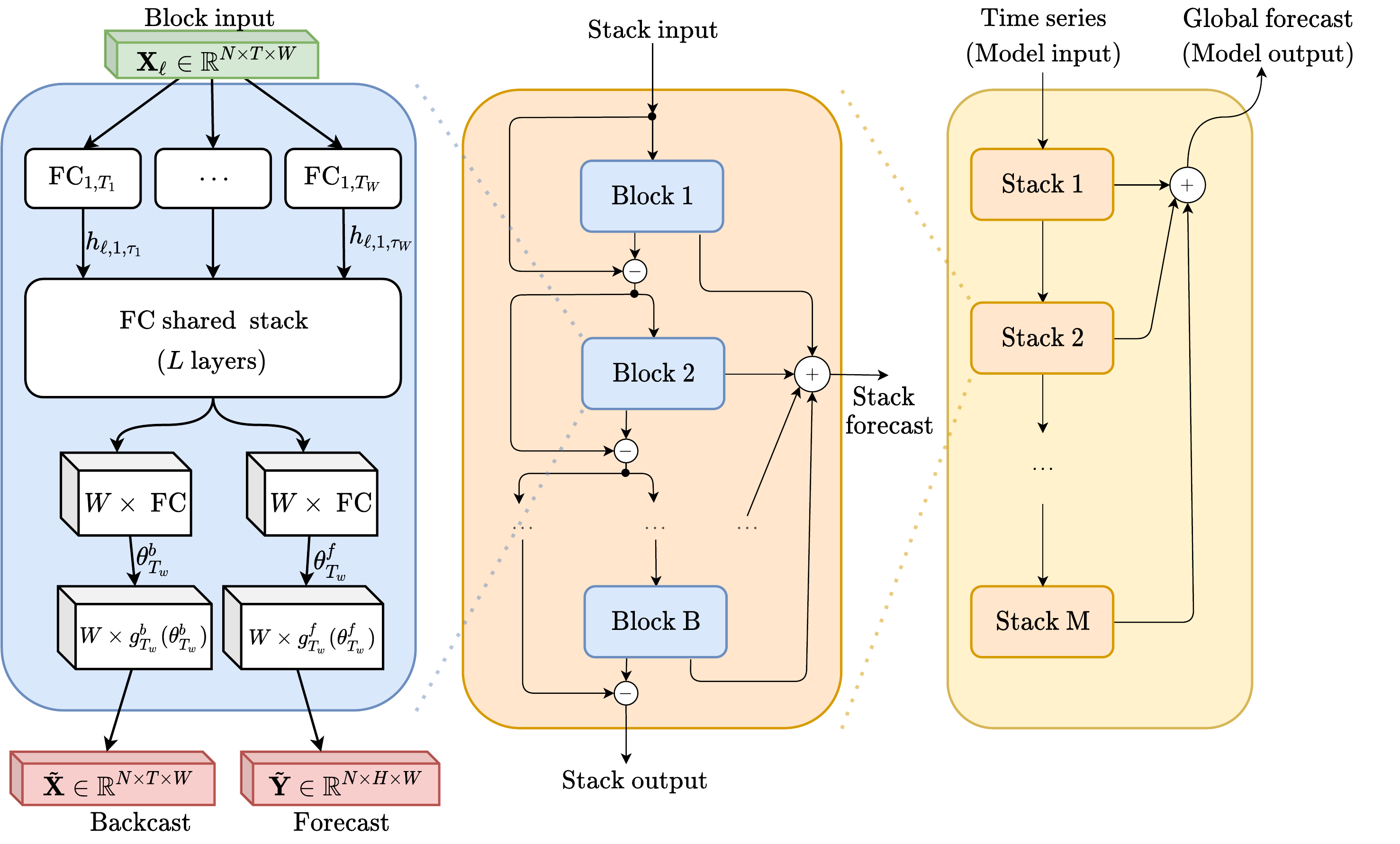}
\caption{Illustration of the proposed model. The basic block consists of multi-head and multi-output fully connected (FC) layers with ReLU non-linear activations, where some layers are shared between the $W$ models. Each block input $\textbf{X}_l \in \mathbb{R}^{N\times L \times W}$ contains the same input signal at different lookback windows $l_1 \cdots l_W$, where for each of the $W$ representations of the signal, missing values are padded with 0. The multi-output part of the block consists of $W$ independent layers (represented by the blue cube in the figure) that predict basis expansion coefficients both forward  $\theta^{f}_{l_w}$ (Forecast) and backward $\theta^{b}_{l_w}$ (Backcast) for each of the $W$ models. A stack can have layers with shared $g^b_{l_w}$ and $g^f_{l_w}$. Forecasts are aggregated by summing over all partial forecasts of each block, enabling us to retrieve which block had the most impact in making the forecast. Parallelization is achieved by forcing head layers of each block to have the same input size, by using mask layers in the input layer to consider only the $T_w$ first observations of input signals and reshaping the tensor to force computation in parallel instead of sequentially applying computation in a loop for each of the $W$ models.}
\label{fig:model}
\end{figure}

The basic building block of the proposed model has a multi-head architecture and is depicted in Fig.~\ref{fig:model} (left). Each $\ell$-th block can take as input up to $W$ input signals $\textbf{x}_{l_w}^{(\ell)}; w\in \{1, \ldots, W\}$ of the same TS with different lookback windows $\boldsymbol{l}=[l_1, \cdots l_W]$, and generates two output vectors for each of the input signals provided: the \textit{backcast} signal $\tilde{\textbf{x}}_{l_w}^{(\ell)}$ of length $l_w$ and the \textit{forecast} signal $\tilde{\textbf{y}}_{l_w}^{(\ell)}$ of length equals to the forecast horizon $H$. We set each $l_w$ to a multiple of $H$ ranging from $2H$ to $7H$.  $\tilde{\textbf{x}}_{l_w}^{(\ell)}$ is fed to the next block for its input and $\tilde{\textbf{y}}_{l_w}^{(\ell)}$ is added to the previous forecast from the previous block. Internally, the basic building block is divided into four parts. 

The first part consists of $W$ independent FC input layers that project the signal into a fixed higher-dimensional representation $\textbf{z}_{l_w}^{(\ell)} \in \mathbb{R}^{+}$. This is done by mapping the $w$-th model with $\phi_w: \mathbb{R}^{l_w} \rightarrow \mathbb{R}^{+}$ such as $\boldsymbol{z}_{l_w}^{(\ell)} = \FC_{l_w}(\boldsymbol{x}_{l_w}^{(\ell)})$. To achieve parallelization in practice, we do this mapping with $\phi_w: \mathbb{R}^{L} \rightarrow \mathbb{R}^{+}$ where $L=max(\boldsymbol{l})$ instead and pad missing values of the $W-1$ signals with 0 that have smaller lookback windows. We set the padding to 0 for the missing values of the lookback and and make sure the FC doesn't have bias to guarantee obtaining the same result as mapping $W$ times the input of each models sequentialy with their respective $\phi_w$.

The second part consists of a shared FC stack that takes as input the TS representation produced in the first part and outputs forward $\theta_{l_w}^{f, (\ell)}$ and backward $\theta_{l_w}^{b, (\ell)}$ predictors of expansion coefficients for each of the $W$ lookback periods.

The third part consists of the $W$ independent backward $g_{l_w}^{b, (\ell)}$ and forward $g_{l_w}^{f, (\ell)}$ basis layers that take as input their respective forward $\theta_{l_w}^{f, (\ell)}$ and backward $\theta_{l_w}^{b, (\ell)}$ expansion coefficients, projecting them over basis functions to produce the backcast $\tilde{\textbf{x}}_{l_w}^{(\ell)} \in \mathbb{R}^{L}$ and the forecast $\tilde{\textbf{y}}_{l_w}^{(\ell)} \in \mathbb{R}^{H}$. 

This approach allows us to parallelize the computation of the forecast by considering an input $\textbf{X}\in \mathbb{R}^{N\times L \times W}$ and producing output $\tilde{\textbf{Y}}\in \mathbb{R}^{N\times H \times W}$ and obtain $W$ forecasts for $N$ TS simulatenously for each of the $W$ lookback windows.

These opperations are repeated iteratively over all blocks across all stacks of the model. Thus, the computation of the forecast and backcast for the $\ell$-th block given the $w$-th signal, is described by the following equations:
\begin{gather}
    \textbf{z}^{(\ell)}_{l_w} = \FC(\FC(\cdots(\FC_{\ell, l_w}(\textbf{x}_{l_w}^{(\ell)})
    \label{eq:block1}
    \\
    \boldsymbol{\theta}_{l_w}^{f, (\ell)} = \FC_{l_w}^{f}(\textbf{z}^{(\ell)}_{l_w})
    \label{eq:block3}
    \\
    \boldsymbol{\theta}_{T_w}^{b, (\ell)} = \FC_{l_w}^b(\textbf{z}^{(\ell)}_{l_w})
    \label{eq:block4}
    \\
    \tilde{\textbf{y}}_{l_w}^{(\ell)} = g_{l_w}^{b, (\ell)}(\boldsymbol{\theta}_{l_w}^{f, (\ell)}) = \sum_{i=1}^{\text{dim}\left(\boldsymbol{\theta}_{l_w}^{f, (\ell)}\right)}\boldsymbol{\theta}_{i,l_w}^{f, (\ell)}\textbf{v}_{i,l_w}^{f, (\ell)}
    \label{eq:block5}
    \\
    \tilde{\textbf{x}}_{l_w}^{(\ell)} = g_{l_w}^{f, (\ell)}(\boldsymbol{\theta}_{T_w}^{b, (\ell)}) = \sum_{i=1}^{\text{dim}\left(\boldsymbol{\theta}_{l_w}^{b, (\ell)}\right)}\boldsymbol{\theta}_{i,T_w}^{b, (\ell)}\textbf{v}_{i,T_w}^{b, (\ell)}
\label{eq:block}
\end{gather}

Here, $\FC$ corresponds to a fully connected layer with ReLU non-linearity activation \cite{nair2010rectified}, and $\textbf{v}_{i,l_w}^{f, (\ell)}$ and $\textbf{v}_{i,l_w}^{b, (\ell)}$ are forecast and backcast basis vectors for the $\ell$-th block. These vectors can either be chosen to be learnable parameters or can be set to specific functional forms that are fixed prior to training the model. In Eq.~\ref{eq:block}, the number of time FC is applied is based on the number of layer and is part of the specification of the model.

Eqs.\ref{eq:block1}-\ref{eq:block} are then repeated iteratively for all blocks, following the same architecture topology as N-BEATS \cite{oreshkin2019n}. The individual blocks are stacked using two residual branches. The first branch, illustrated in Fig.~\ref{fig:model} (middle), runs over the backcast signal produced by each block and iteratively decomposes the initial TS signal such that the subsequent block consider the residual of its preceding block. The second branch, illustrated in Fig.~\ref{fig:model} (right), aggregates the partial forecast of each block. These operations are described by the following equations:
\begin{equation}
    \textbf{x}_{l_w}^{(\ell+1)} = \textbf{x}_{l_w}^{(\ell)} - \tilde{\textbf{x}}_{l_w}^{(\ell)}
\end{equation}
\begin{equation}
    \tilde{\textbf{y}}_{l_w} = \sum_{\ell}\tilde{\textbf{y}}_{l_w}^{(\ell)}
\end{equation}

\subsection{Generic and Interpretable Model Version}
\label{sec:Generic and Interpretable version}

Multiple versions of this approach can be provided to parameterize each of the $W$ models. For instance, both the generic and interpretable versions of N-BEATS proposed in \cite{oreshkin2019n} are compatible with our model. We will briefly describe these two extensions; we refer the reader to the original paper for more details \cite{oreshkin2019n}.

\textbf{The generic architecture:} in this version, $g_{l_w}^{b}$ and $g_{l_w}^{f}$ are specified as a linear projection of the previous layer output such that the outputs of the $\ell$-th block are described as follows:
\begin{equation}
    \tilde{\textbf{y}}_{l_w}^{(\ell)} = \textbf{V}_{l_w}^{f, (\ell)}\boldsymbol{\theta}_{l_w}^{f, (\ell)} + \textbf{B}_{l_w}^{f, (\ell)} \quad \tilde{\textbf{x}}_{l_w}^{(\ell)} = \textbf{V}_{l_w}^{b, (\ell)}\boldsymbol{\theta}_{l_w}^{b, (\ell)} + \textbf{B}_{l_w}^{b, (\ell)}
\end{equation}
where $\textbf{V}_{l_w}^{f, (\ell)} \in \mathbb{R}^{H\times dim\left(\boldsymbol{\theta}_{l_w}^{f, (\ell)}\right)}$, $\textbf{B}_{l_w}^{b, (\ell)} \in \mathbb{R}^{H}$ and $\textbf{V}_{l_w}^{b, (\ell)} \in \mathbb{R}^{L\times dim\left(\boldsymbol{\theta}_{l_w}^{b, (\ell)}\right)}$, $\textbf{B}_{l_w}^{b, (\ell)} \in \mathbb{R}^{L}$ are basis vectors learned by the model, which can be taught as waveforms.  Because no additional constraints are imposed on the form of $\textbf{V}_{l_w}^{f, (\ell)}$ the waveforms learned do not have inherent
structure on how they should look.

\textbf{The interpretable architecture:} Similar to the traditional TS decomposition into trend and seasonality found in \cite{cleveland1990stl, us2016x}, trend and seasonality decomposition can be enforced in $\textbf{V}_{l_w}^{f, (\ell)}$ and $\textbf{V}_{l_w}^{b, (\ell)}$. \cite{oreshkin2019n} proposed to do this by conceptually separating the set blocks into two stacks such that one stack of blocks is parameterized with a \textbf{trend model} ($\boldsymbol{T}$) and the other with a \textbf{seasonal model} ($\boldsymbol{S}$). All block in a stack shared the same parameters. The \textbf{trend model} consists of constraining the basis function to modelize a trend signal, i.e., using a function polynomial of small degree $p$ as follows:
\begin{equation}
   \tilde{\textbf{y}}_{l_w}^{(\ell)}=g_{l_w,  \textrm{trend}}^{f, (\ell)}(\boldsymbol{\theta}_{l_w}^{f, (\ell)})=\textbf{T}\boldsymbol{\theta}_{l_w}^{f, (\ell)}; \textbf{T}=[\textbf{1},t,\cdots t^{p}]
\end{equation}
where $\textbf{T}$ is a matrix of powers of $p$. Thus the waveform extracted will follow a monotonic or a slowly varying function. 
The \textbf{seasonal model} constrains the basis function to modelize periodic functions, i.e, $g_{T_w}^{f, (\ell)}(\boldsymbol{\theta}_{l_w}^{f, (\ell)};\textbf{V}_{t, l_w}^{f, (\ell)})$, using Fourier series as follows:
\begin{gather}
    \tilde{\textbf{y}}_{l_w}^{(\ell)}=g_{l_w,  \textrm{seas.}}^{f, (\ell)}(\boldsymbol{\theta}_{l_w}^{f, (\ell)})=\textbf{S}\boldsymbol{\theta}_{l_w}^{f, (\ell)};\\\textbf{S}=[\textbf{1}, cos(2\pi\textbf{t}, \cdots , cos(2\pi \lfloor{H/2-1}\rfloor\textbf{t}),sin(2\pi \lfloor{H/2-1}\rfloor\textbf{t})] \nonumber
\end{gather}
Thus, by first (1) applying the \textbf{trend model} and then (2) applying the \textbf{seasonal model} within the doubly residual stacking topology of the model, we obtain a model that applies TS component decomposition in a similar way to than traditional decomposition approaches. Basis functions are a generalization of linear regression where we replace each input with a function of the input. Here, the polynomial and the Fourier series are functions that model uses the trend and seasonality and take as input the embedding computed from the TS at each block ant not the raw TS values. In the case of the generic version, the basis functions for the forecasts and the backast are respectively the vectors $ \textbf{V}_{l_w}^{f, (\ell)}$ and $ \textbf{V}_{l_w}^{b, (\ell)}$.

In any configuration of the model, estimating the parameters of the model problem is done by maximum likelihood estimation (MLE). To simplify the notation, we consider eq.~\ref{eq:simplenbeats} as the function that establishes the forecast, where $\boldsymbol{\theta}_{\NBEATS}$ is the set of all parameters of each block and $\textbf{x}^{i}_{l_w}$ is the $i$-th TS considered with input size of length $l_w$.
\begin{equation}
    \tilde{\textbf{y}}_{l_w}Y{i} = \NBEATS(\textbf{x}^{i}_{l_w}; \boldsymbol{\theta}_{\NBEATS})
\label{eq:simplenbeats}
\end{equation}
Thus, optimizing the model consists of optimizing eq.~\ref{eq:loss_NBEATS}. We use a stochastic gradient descent optimization with Adam \cite{kingma2014adam} over a fixed set of itterations and a three-steps learning rate schedule. Here $\mathcal{L}(NBEATS(\textbf{x}^{n}_{T_w}; \theta_{\NBEATS}), \textbf{y}^{(n)})$ corresponds to some metric function that measures the quality of the forecast to the ground truth $\textbf{Y}$. Note that we combine the losses of the forecasts of all models, using the mean values to promote cooperation between the different models. Following the same training framework as \cite{oreshkin2019n}, we used the MAPE, MASE and SMAPE losses to build the ensemble, all of which are detailed in the following section. We refer the reader to \cite{oreshkin2019n} for design choice of the model and a exhaustive discussion on the parameter choice of this model. In our work we reuse the same set of parameters and do not apply hyper-parameters search at the exception of the yearly TS where our model converge to a stable results earlier (10k itterations instead of 15k).

\begin{equation}
\label{eq:loss_NBEATS}
 \theta^{*}_{\NBEATS}= \operatorname*{argmin}_{\theta^{*}_{\NBEATS}} {\dfrac{1}{N}}\sum_{i=0}^{N} \frac{1}{W}\sum_{w=1}^{W}\mathcal{L}(\NBEATS(\textbf{x}^{i}_{l_w}; \theta_{\NBEATS}), \textbf{y}^{i})
\end{equation}

\section{Experimental setup}
\label{sec:Experimental analysis}

We conducted the experimental evaluation of the forecasting methods on 6 datasets which include a total of 105'968 unique TS when combined and over 2.5 million forecasts to produce on these TS. We report the accuracy of our model on the first and dataset and consider the rest to assess the model ability to generalize in other settings. We details all datasets here and report our generalization results on zero-shot forecasting in \ref{sec:zero-shot forecasting}. The datasets are the following:
\begin{enumerate}[label={(\arabic*}),leftmargin=*]
    \item (\textit{public}) \textbf{M4:} 100'000 heterogeneous TS from multiple sectors that include economic, finance, demographics and other industry used in the M4 TS competition \cite{makridakis2018m4,makridakis2020m4}. 
    \item (\textit{public}) \textbf{M3:} 3003 heterogeneous TS from derived from mostly from financial and economic domains \cite{koning2005m3}.
    \item (\textit{public}) \textbf{Tourism:} 1311 TS of indicators related to tourism activities sampled monthly, quarterly and yearly \cite{athanasopoulos2011tourism,athanasopoulos2011value}.
    \item (\textit{public}) \textbf{Electricity:} 370 TS of the hourly electricity usage of 370 customers over three years \cite{Dua:2019,yu2016temporal}.
    \item (\textit{public}) \textbf{Traffic:} 963 TS of the hourly occupancy rates on the San Francisco Bay Area freeways scaled between 0 and 1 \cite{Dua:2019,yu2016temporal}.
    \item (\textit{proprietary}) \textbf{Finance:} 321 TS observed between 2005-07-01 and 2020-10-16 of the adjusted daily closing price of various U.S. mutual funds and exchange traded funds traded on U.S. financial markets, each covering different types of asset classes including stocks, bonds, commodities, currencies and market indexes, or a proxy for a market index covering a larger set of financial asset than the dataset used in \cite{chatigny2021spatiotemporal}.
\end{enumerate}
For the \textbf{M4, M3} and \textbf{Tourism} datasets, target TS trajectories were specified by the competition’s organizers with each subpopulation of TS with the same frequency (Hourly, Quarterly, etc..) having its own horizon (6, 8, etc...). For the \textbf{Electricity}  and \textbf{Traffic} datasets, the test was set using rolling window operation as described in \ref{sec:Electricity and Traffic Datasets Details}. For the \textbf{Finance} dataset, the forecast was evaluated on three rolling forecast setups by sampling the TS on different frequencies, i.e.: daily, weekly and monthly. In total there are 2'602'878 individual TS that were sampled from the 321 original ones across 3 forecast horizons. Despite the dataset being collected from proprietary data sources which we cannot redistribute, we provide the necessary details to help interested readers reconstruct the datasets in \ref{sec:Finance dataset}. A summary of the statistical properties, forecast horizons and metadata of these dataset are presented in ~\ref{sec:Dataset}.

We trained our model on the \textbf{M4} dataset on the TS each subpopulations, i.e. [Yearly, Quarterly, Monthly, Weekly, Daily, Hourly]. We replicated the results from \cite{oreshkin2019n} by training the two N-BEAT model variants discussed in Sec.~\ref{sec:Generic and Interpretable version} using the implementation provided by the original authors and with scaled TS where we divided all TS observations by the maximum values observed. This scaling was done per TS with respect to the lookback window. For our model, the scaling was done on by dividing all lookback windows by the maximum value observed over all lookback windows.

We compared the forecast accuracy of our approaches with the reported accuracy of other TS models in the M4 TS competition including FFORMA \cite{montero2020fforma} and ES-RNN \cite{smyl_hybrid_2020}. In reporting the accuracy of these models, we relied upon the accuracy and the pre-computed forecasts reported in their respective original paper. The statistical models were produced on R using the forecast package \cite{hyndman2008automatic} and we measured the training time to train and produce each forecast of our model as well as the Theta method. We also relied upon the reported running time of the implementation provided in \cite{makridakis2020m4}. Finally, all models were compared on a naive forecast, i.e., a random walk model or a seasonally adjusted random walk, that assumes all future values will be the same as the last known one(s). This was done to assess whether the forecasts of these models are accurate in the first place.

\begin{equation}
\label{eq:MAPE}
MAPE(\boldsymbol{\tilde{x}}, \boldsymbol{x}) = \frac{100}{H}\sum_{i=1}^{H}\frac{|\boldsymbol{\tilde{x}}_{T+i}-\boldsymbol{x}_{T+i}|}{\boldsymbol{x}_{T+i}}
\end{equation}
\begin{equation}
\label{eq:MASE}
MASE(\boldsymbol{\tilde{x}}, \boldsymbol{x}) = \frac{1}{H}\frac{\sum_{i=1}^{H}|\boldsymbol{x}_{T+i} - \boldsymbol{\tilde{x}}_{T+i}|}{\frac{1}{T-m}\sum_{t=m+1}^{T}|\boldsymbol{x}_t - \boldsymbol{x}_{t-m}|}
\end{equation}

\begin{equation}
\label{eq:SMAPE}
    SMAPE(\boldsymbol{\tilde{x}}, \boldsymbol{x}) = \frac{200}{H}\sum_{i=1}^{H}\frac{|\boldsymbol{x}_{T+i} - \boldsymbol{\tilde{x}}_{T+i}|}{|\boldsymbol{x}_j| + |\boldsymbol{\tilde{x}}_{T+i}|}
\end{equation}
\begin{equation}
\label{eq:OWA}
OWA(\boldsymbol{\tilde{x}}, \boldsymbol{x}) = \frac{1}{2} \left[
\frac{SMAPE}{SMAPE_{NAIVE2}} + 
\frac{MASE}{MASE_{NAIVE2}}
\right]
\end{equation}

\begin{equation}
\label{eq:ND}
    ND(\boldsymbol{\tilde{x}}, \boldsymbol{x}) = \frac{\sum_{i=1}^{H}|\boldsymbol{\tilde{x}}_{T+i}-\boldsymbol{x}_{T+i}|}{\sum_{i=1}^{H}|\boldsymbol{x}_{T+i}|}
\end{equation}

\begin{equation}
\label{eq:MDA}
    MDA(\boldsymbol{\tilde{x}}, \boldsymbol{x}) = \frac{1}{H}\sum_{i=1}^H \text{sign}(\boldsymbol{\tilde{x}}_{T+i}-\boldsymbol{x}_{T})=\text{sign}(\boldsymbol{x}_{T+i}-\boldsymbol{x}_{T})
\end{equation}


We evaluated the forecast accuracy using 8 standard TS metrics: the mean absolute percentage error (MAPE) used in the Tourism compeition \cite{athanasopoulos2011tourism}, the mean absolute scaled error (MASE) \cite{hyndman2006another}, the scaled mean absolute percentage error (SMAPE) used in the M3 competition \cite{koning2005m3}, the normalized deviation (ND) used in \cite{salinas2020deepar} and the mean directional accuracy (MDA). Additionally for the M4 competition, we evaluated the model on the overall weighted average (OWA) between the SMAPE and the MASE such that a seasonally-adjusted naive (NAIVE2) forecasting model obtains a score of 1.0 \cite{makridakis2020m4}. For instance, an OWA of 0.90 means that the forecast is on average 10\% better than a NAIVE2 model with respect to both the SMAPE and MASE metrics. The MDA measures the model's ability to produce forecasts where the trajectory follows the actual change of the TS relative to the last known value: the higher the MDA is, the better a model predicts the trend of a TS at any given time. For all other metrics, the lower the value, the better a model predicts the TS. For the \textbf{M4} dataset, we only consider the OWA, the MASE and the MDA. The other metrics were used in order to compare ourselve with other methods and other datasets as detailed in \ref{sec:zero-shot forecasting}.

Eq.~\ref{eq:MAPE}-\ref{eq:MDA} describes how these metrics are computed. $\tilde{x}$ is the forecast, $x$ is the ground truth and $m$ is the time interval between successive observations considered by the organizers for each data frequency, i.e., 12 for monthly, four for quarterly, 24 for hourly and one for yearly, weekly and daily data. Without loss of generality to previous equatios $T$ is the number of point in-sample observed to make the forecastushe and $H$ is the forecast horizon.

\subsection{Baseline and Benchmark}
\label{sec:Baseline and benchmark}

\begin{figure}[htbp]
\centering
\begin{minipage}{.5\textwidth}
  \centering
  \includegraphics[width=\linewidth]{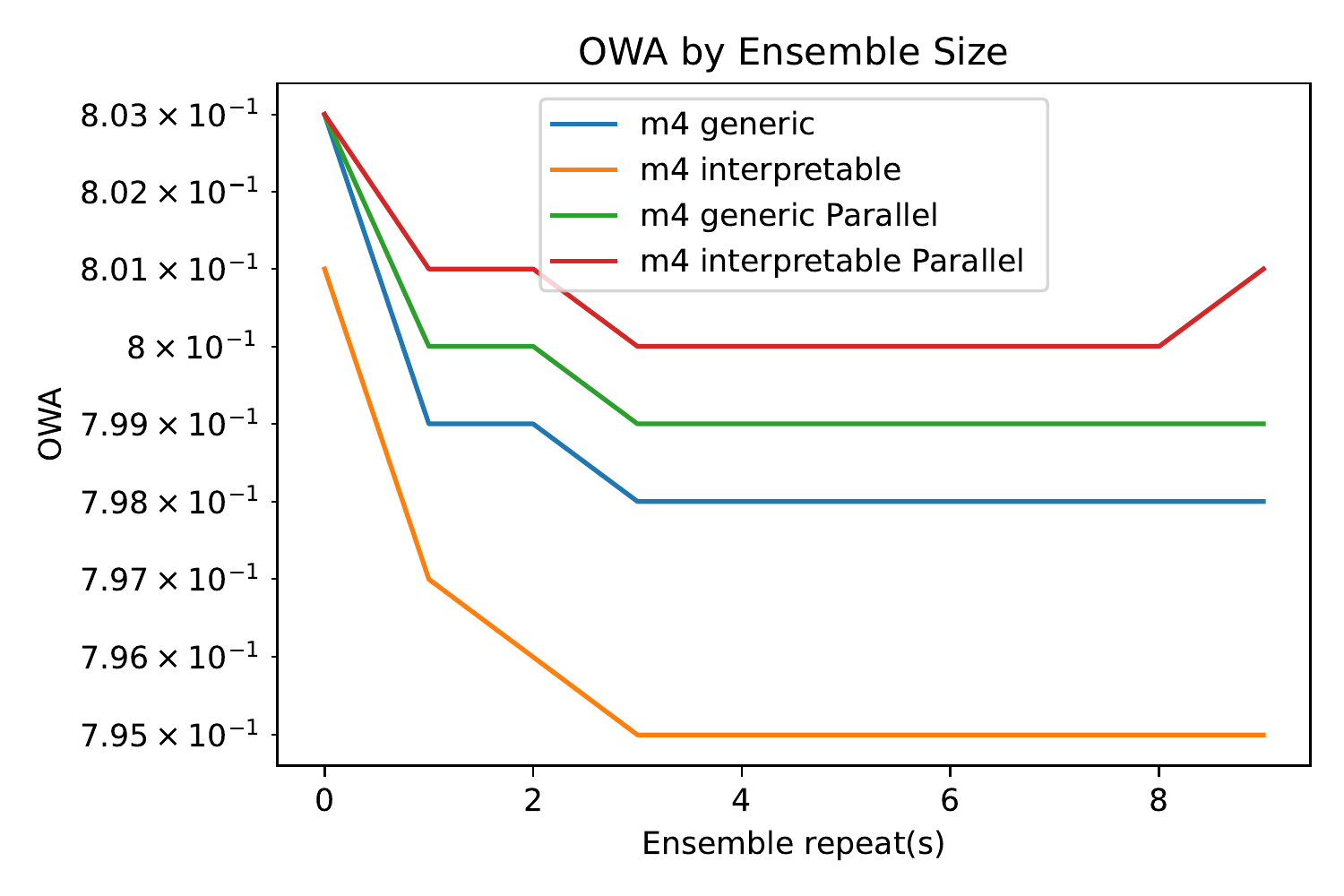}
\end{minipage}%
\begin{minipage}{.5\textwidth}
  \centering
  \includegraphics[width=\linewidth]{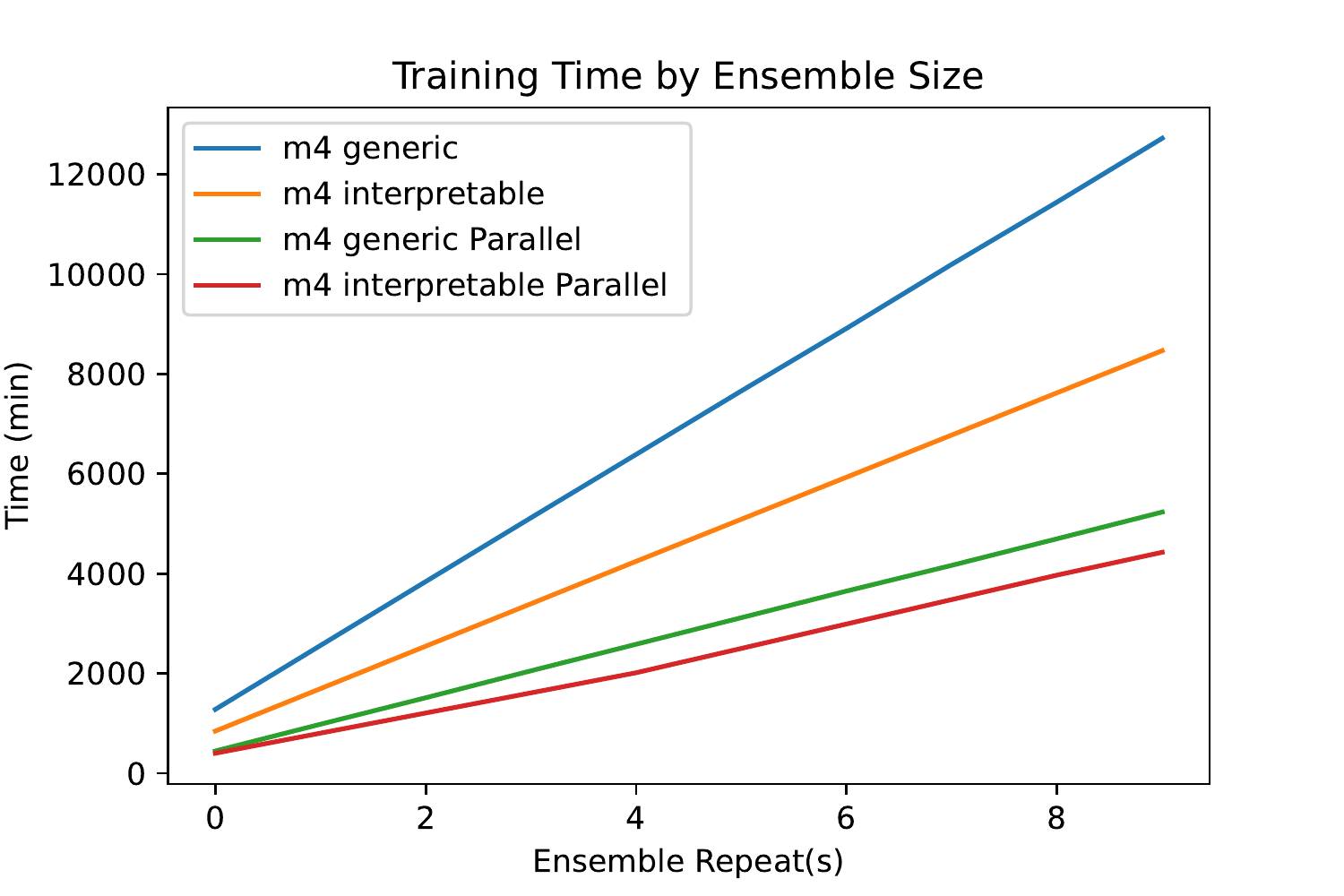}
\end{minipage}.
\caption{OWA metric (left) and Time (right) in minutes for the different N-BEATS models configurations as a function of ensemble size.}
\label{fig:NBEATS_scaling_comparison}
\end{figure}

We present the results of the baseline and benchmark accuracies for the M4 dataset in Table~\ref{Tab:baseline_accuracy}. The table gives the reported accuracy of N-BEATS reported in the original papers~\cite{oreshkin2019n}, the replicated results using the publicly accessible implementation provided by the original authors along with their scaled versions \cite{oreshkin2020meta} based upon their implementation and our model NBEATS(P). Three main conclusions can be drawn:
\begin{enumerate}[label={\textbf{(\arabic*})}]
\item Scaling TS to allow generalization on other datasets for the N-BEATS model, as presented in \cite{oreshkin2020meta}, adds a penalty on the OWA metrics for the M4 dataset, which suggests that there is a trade-off between accuracy and generalization on other datasets for DNN-based models.

\item Figure~\ref{fig:NBEATS_scaling_comparison} details  how sensemble size has an impact on computational time to train. It can be seen that applying a bagging procedure \cite{breiman1996bagging} 3 to 4 times is sufficient to get an accurate ensemble for both the NBEATS and NBEATS(P) model but NBEATS(P) is more efficient the larger the ensemble size is. 

\item The top-performing models do not differ significantly with respect to the coverage of the TS forecasted and the mean directional accuracy (MDA). This provides an argument that if one is mainly interested in predicting the TS variation from the forecast origin, relying on the fastest implementation of the top-performing models for a first initial prediction is a cost-effective solution. 

\end{enumerate}

\begin{figure}[htbp]
\centering
\begin{minipage}{.5\textwidth}
  \centering
  \includegraphics[width=\linewidth]{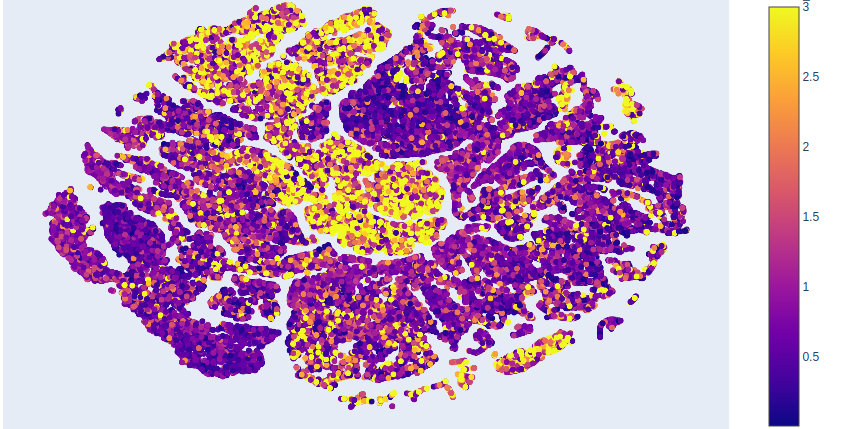}
\end{minipage}%
\begin{minipage}{.5\textwidth}
  \centering
  \includegraphics[width=\linewidth]{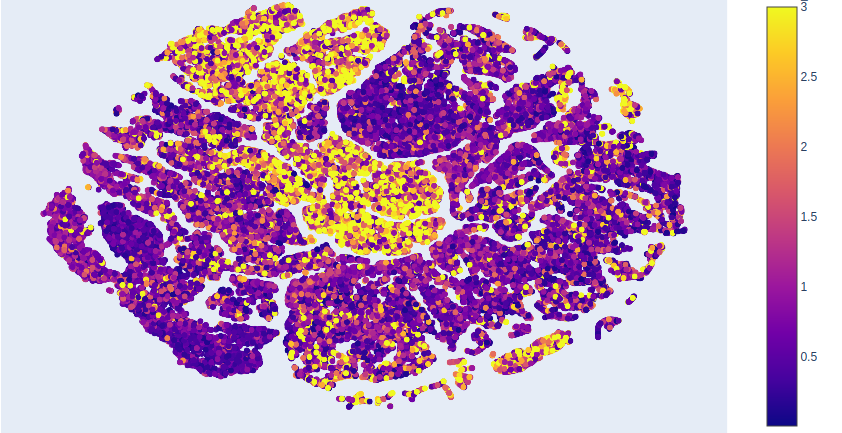}
\end{minipage}.
\captionof{figure}{MASE coverage for the ES-RNN (left) and N-BEATS(i) (right) over the M4 dataset. Each point on the graphs corresponds to a single TS and the darker its color, the better the given model according to the MASE. Horizontal and vertical coordinates represent the value of the two-dimensional embedding computed with TSNE~\cite{van2008visualizing} from the statistical features of the series which we detail in \ref{sec:Dataset}. All TS with MASE values over 3 where assigned the same color to facilitate visualization.}
\label{fig:tsnee_mase_comparison}
\end{figure}

Regarding \textbf{(2)}, we illustrate this phenomenon in fig.~\ref{fig:tsnee_mase_comparison} by plotting the TSNE embedding of each series of M4 computed from the same set of features used in the FFORMA \cite{montero2020fforma} by comparing the top performing model with the N-BEATS(i) model by coloring each TS with its individual MASE accuracy. We refer the reader to table.\ref{tab:features_detail} and \cite{montero2020fforma, hyndman2019tsfeatures} for a detailed overview of the 42 features used and their interpretation. \ref{sec:Dataset} details the distribution of these features over all datasets we considered. Note that there are no substantial differences between the approaches, despite some subtle regions of the graph where we can observe N-BEATS(i) performing better overall than ES-RNN.

\begin{table}[htbp]
\begin{adjustbox}{max width=\textwidth}
    \begin{tabular}{lrrrrrrrr}\toprule
        &\multicolumn{1}{c}{\textbf{Yearly}}&\multicolumn{1}{c}{\textbf{Quarterly}}&\multicolumn{1}{c}{\textbf{Monthly}}&\multicolumn{1}{c}{\textbf{Others}}&\multicolumn{1}{c}{\textbf{Average}}&\multicolumn{1}{c}{\textbf{Coverage}}
        \\\cmidrule(r){2-7}   
        &\multicolumn{5}{c}{OWA}&\multicolumn{1}{c}{MASE \% ($< T=1.0$)}&\multicolumn{1}{c}{MDA(\%)}
        \\\midrule
        \textit{NAIVE}  & 1.000 & 1.066 & 1.095 & 1.335 & 1.058 & 40.299 & 3.2 
        \\
        \textit{NAIVE2} & 1.000 & 1.000 & 1.000 & 1.000 & 1.000 & 43.288 & 33.1 
        \\
        \textit{SNAIVE} & 1.000 & 1.153 & 1.146 & 0.945 & 1.078 & 36.095 & 42.7 
        \\
        \midrule
        \textit{ARIMA} \cite{box2015time} & 0.892 & 0.898 & 0.903 & 0.967 & 0.903 & 51.145 & 53.8 
        \\
        \textit{HOLT} \cite{chatfield1991prediction} & 0.947 & 0.932 & 0.988 & 1.180 & 0.971 & 48.659 & 61.7 
        \\
        \textit{ETS} \cite{hyndman2002state} & 0.903 & 0.890 & 0.914 & 0.974 & 0.908 & 50.987 & 48.6 
        \\
        \textit{THETA} \cite{assimakopoulos2000theta} & 0.872 & 0.917 & 0.907 & 0.995 & 0.897 & 48.686 & 61.7 
        \\
        \textit{SES} \cite{hyndman2008forecasting} & 1.002 & 0.970 & 0.951 & 0.995 & 0.975 & 44.719 & 35.3 
        \\
        \textit{DAMPED} \cite{makridakis2020m4} & 0.890 & 0.893 & 0.924 & 1.005 & 0.907 & 49.838 & 61.1 
        \\
        \textit{COMB} \cite{makridakis2020m4} & 0.867 & 0.890 & 0.920 & 1.039 & 0.898 & 49.784 & 61.3 
        \\
        \midrule    
        \textit{MLP}' \cite{makridakis2020m4, goodfellow2016deep} & 1.288 & 1.684 & 1.749 & 3.028 & 1.642 & 26.603 & 60.6 
        \\
        \textit{RNN}' \cite{makridakis2020m4, goodfellow2016deep} & 1.308 & 1.508 & 1.587 & 1.702 & 1.482 & 28.437 & 59.8 
        \\
        \midrule
        \textit{ProLogistica} \cite{pawlikowski2020weighted} & 0.820 & 0.855 & 0.867 & 0.742 & 0.841 & 53.620 & 62.6 
        \\
        \textit{FFORMA} \cite{montero2020fforma} & 0.799 & 0.847 & 0.858 & 0.914 & 0.838 & 53.418 & 63.7 
        \\
        \textit{ES-RNN} \cite{smyl_hybrid_2020} & 0.778 & 0.847 & 0.836 & 0.920 & 0.821 & 53.271 & 63.2 
        \\
        \textit{N-BEATS (I)} \cite{oreshkin2019n} & 0.765 & 0.800 & 0.820 & 0.822 & 0.797 & --- & ---  
        \\
        \textit{N-BEATS (G)} \cite{oreshkin2019n}& 0.758 & 0.807 & 0.824 & 0.849 & 0.798 & --- & ---  
        \\
        \textit{N-BEATS (I+G)} \cite{oreshkin2019n} & 0.758 & 0.800 & 0.819 & 0.840 & 0.795 & --- & --- 
        \\
        \midrule
        \textbf{Ours:}\\
        N-BEATS (G) \cite{oreshkin2019n} & 0.770 & 0.793 & 0.818 & 0.832 & 0.798 & 55.576 & 64.6 
        \\
        N-BEATS (I) \cite{oreshkin2019n} & 0.763 & 0.797 & 0.817 & 0.838 & 0.795 & 55.600 & 63.7 
        \\
        N-BEATS (I+G) \cite{oreshkin2019n} & 0.761 & 0.792 & 0.814 & 0.834 & 0.793 & 55.868 & 64.6 
        \\
        \midrule
        N-BEATS (G) \textit{scaled} \cite{oreshkin2020meta} & 0.784 & 0.810 & 0.827 & 0.836 & 0.809 & 54.960 & 64.5 
        \\
        N-BEATS (I) \textit{scaled} \cite{oreshkin2020meta} &0.773 & 0.817 & 0.826 & 0.843 & 0.806 & 54.919 & 63.7 
        \\
        N-BEATS (I+G) \textit{scaled} \cite{oreshkin2020meta} & 0.778 & 0.814 & 0.824 & 0.836 & 0.806 & 55.109 & 64.4 
        \\ 
        \midrule
        \textbf{N-BEATS parallel (G)} & 0.764 & 0.804 & 0.820 & 0.855 & 0.799 & 55.332 & 64.4 
        \\
        \textbf{N-BEATS parallel (I)} & 0.759 & 0.817 & 0.824  & 0.850 & 0.801 & 54.966 & 63.7 
        \\
        \textbf{N-BEATS parallel (I+G)} & 0.757 & 0.806 & 0.820 & 0.851 & 0.796 & 55.375 & 64.5 
        \\
        \bottomrule
        \textbf{N-BEATS parallel (G) \textit{scaled}} & 0.775 & 0.829 & 0.833 & 0.851 & 0.812 & 54.506 & 63.8 
        \\
        \textbf{N-BEATS parallel (I) \textit{scaled}} & 0.772 & 0.845 & 0.844 & 0.867 & 0.819 & 53.772 & 63.6 
        \\
        \textbf{N-BEATS parallel (I+G) \textit{scaled}} & 0.771 & 0.834 & 0.835 & 0.854 & 0.813 & 54.344 & 63.9 
        \\
        \bottomrule
    \end{tabular}
    \caption{Averaged forecasting results of the M4 competition for the evaluated models. The OWA metric is presented for each seasonal pattern observed. Forecasts from models in \textit{italics} were pre-computed except for the N-BEATS models. We replicate the results with the implementation provided by the authors, e.g: \textit{N-BEATS (I)} \textbf{(original)} vs N-BEATS (I) \textbf{(our)}. \textit{MLP}' and \textit{RNN}' models are appended with "'" to signify that these model were trained per TS using a seasonal and trend decomposition with manual pre- and post processing steps \cite{makridakis2020m4}. We also considered a coverage indicator which measures the number of series that a model forecasts better than an arbitrary MASE accuracy threshold of $\tau=1.0$. We also added the MDA of the forecast.}
    \label{Tab:baseline_accuracy}
\end{adjustbox}
\end{table}

\subsection{Training Time and Number of Parameters}
\label{sec:Training time and number of parameters}

Table~\ref{tab:time_and_corr} presents the time to train each model accuracy as well as the average pairwise absolute percentage error correlation of the forecast residuals between ensemble members, in a way to similar to the experimental evaluation of M4 submission performed in \cite{agathangelou2020correlation}. At first glance, we see that the training time of these models be as long as multiple weeks. N-BEATS models timing are reported with a single GPU. In comparison to N-BEATS, N-BEATS(P) takes less time regardless of the variant used but the gain is observed especially for the generic architecture. Both the original approach and ours are at the same level of correlation and, while ours is slightly less diverse, it is roughly twice as efficient and achieves the same level of accuracy. We can observe that using the scaled version has little impact in terms of diversity. In our preliminary result, we also observed that there was no significant difference in terms of the TS samplers used to train N-BEATS(P) where for instance, different TS were sampled for the $W$ model. 

In practice, computational time remains more than significant for training these models on a single GPU. However, we can speed up the training by training model simulatenously. If we consider N-BEATS(G) vs N-BEATS(P, G), each ensemble member trained on a single TS frequency takes on average 12 \& 27 minutes, and at worst, 19 \& 57 minutes respectively. N-BEATS(G)'s time is for a single lookback and thus requires 1080 ($6\text{ lookback}\times3\text{ losses}\times6\text{ frequencies}\times10\text{ repeats}$) independent models to be trained whereas N-BEATS(P, G) requires only ($3\text{ losses}\times6\text{ frequencies}\times10 \text{ repeats}$) independent models. Thus, if one would have access to 1080 GPUS, the total training time of N-BEATS(G) could be done in 20 min, but this is an unrealistic amount of ressources for most organization. Our approach cuts down that cost: given 10 GPUs, the N-BEATS(G) and a greedy schedudling of model training, it would take roughly 35h to train whereas our would only take 16h with all 10 repeats. Using 20 gpus, our model would achieve it in 8 hours whereas the previous model would take 18. 

When forecasting larger amount of TS, say 1 billions monthly distinct TS, estimating the cost can be difficult. We can make a reasonable assumption that the computational time required to train a model scales linearly with the number of time series to forecast altought it takes roughly the same to train on different subpopulations or the other based on the number of itterations. Hence assuming we are using the same 3 losses and bagging procedure and it takes a single model to train the monthly TS where the number of itterations required linearly scaled from the one used in M4 (see \ref{tab:m4_hpCNN}) and require 75k itterations instead, it would take 1878h and 1551h for the generic and interetable version of our model on 30 GPUS. Altough a larger dataset may benefit from a deeper \& wider model further inflating the cost, the number of itterations might not need to be this high. However, our approach will result in similar gain in the scenario of deeper \& wider model. Regardless, training these model for everyday usage requires a lot of computational ressouce. In order for the cost to be kept low at this scale, training would need to be done less frequently and models would have to remain outdated to some extent as recent trends and structural changes in the data wouldn't be used to update the model parameters. 

To further show the performance of our model, we show in Table.~\ref{tab:smaler_ensemble} that one could have achieved the same average accuracy as the top M4 competitions entries by training our for approximatively 2h on a single GPU without bagging. Thus, even with minimal amount of ressources, smaller organizations can train 54 DNN-based models, each on TS of different freqencies, losses and lookback windows very fast making our model far more accessible to small organization who doesn't have dozens of gpus available.

Since the training procedure of our approach takes roughly a fixed amount of time to train regardless of the number of TS to forecast\footnote{This is because the procedure to train our model is itteration-based and not epoch-based. The term "epochs" refers to the number of passes of \emph{the entire training dataset} our models has seen. Our approach differs in that we itterate on batches of TS sampled and sliced randomly at different cut-off points.}, forecasting more TS might requires more itterations and/or more parameters for the model to capture the dynamics of these additonal TS. Therefore the training time is expected to increase the more TS we want to forecast in the training regime. However, the overall training time doesn't increase linearly with the number of TS to forecast as increasing the number of itteration is a fixed cost and the number of itterations to train Quarterly (24K TS) or Monthly (48K TS) was the same in our setting. This leaves, the total number of itterations to train all models, the forecast horizons, the number of parameters and the number of models trained simultanously to have effect on training time. 

The difference in improvement factor between parallelized generic and interpretable versions of N-BEATS(P) is due to the hidden layer sizes between the two versions. Having a higher number of hidden neurons reduce the computational gain of training multiple models conjointly as it saturate GPU usage. If we have a sufficiently expressive model without requiring too many hidden neurons, N-BEATS(P) is expected to produce accurate forecasts at a fraction of the cost. Otherwise the gain will be diminished. Regardless, these results show that ensemble diversity and accurate forecasts could have been achieved with reduction in resources and computation time.

\begin{table}[htbp]
\begin{adjustbox}{max width=\textwidth}
    \begin{tabular}{lrr|lrr}\toprule
        &\multicolumn{1}{c}{\textbf{Time (min.)}}&\multicolumn{1}{c}{\textbf{Corr. (mean, std.)}}&&\multicolumn{1}{c}{\textbf{Time (min.)}}&\multicolumn{1}{c}{\textbf{Corr. (mean, std.)}}
        \\\midrule
        Theta & 7.28 & $-- \pm --$ & \multicolumn{3}{c}{ }\\\midrule
        ProLogistica \cite{pawlikowski2020weighted} & 39655 & $-- \pm --$ & \multicolumn{3}{c}{ }\\
        FFORMA \cite{montero2020fforma}: & 46108 & $-- \pm --$ & \multicolumn{3}{c}{ }\\
        ES-RNN \cite{smyl_hybrid_2020}: & 8056 & $-- \pm --$ & \multicolumn{3}{c}{ }\\
        N-BEATS(G) \cite{oreshkin2019n} & 11773 & $0.85\pm 0.02$ & N-BEATS(G) \textit{scaled} &  11755 & $0.84\pm 0.04$\\
        N-BEATS(I) \cite{oreshkin2019n} & 7437 & $0.89\pm 0.02$
        & N-BEATS(I) \textit{scaled} & 6607 & $0.85\pm 0.03$\\
        N-BEASTS(I+G) \cite{oreshkin2019n} & 19211 & $0.84\pm 0.03$ & N-BEATS(I+G) \textit{scaled} &  19170 & $0.84\pm 0.03$ \\
        \midrule
        N-BEATS(P, G)   \textbf{(our)}& \textbf{5301}
        & $0.87\pm 0.02$ & N-BEATS(P,G)  \textit{scaled}   \textbf{(our)}& \textbf{6157} & $0.90\pm 0.02$ \\
        N-BEATS(P, I)   \textbf{(our)}& \textbf{6990} & $0.89\pm 0.03$ & N-BEATS(P,I)  \textit{scaled}  \textbf{(our)}& \textbf{4785} & $0.89\pm 0.08$ \\
        N-BEATS(P, I+G)   \textbf{(our)}& 11840 & $0.88\pm 0.02$ & N-BEATS(P,I+G)  \textit{scaled}  \textbf{(our)}& 10943 & $0.89\pm 0.06$\\
        \bottomrule
    \end{tabular}
    \caption{Time required to train to train all members of the ensemble of our models vs other and average \& standard deviation of the absolute percentage correlation between ensemble members on the test sets. We include the total time to produce a forecast for the theta method for comparison. Except for Prologistica, FFORMA and ES-RNN whose training time was replicated in \cite{makridakis2020m4}, the total time presented is with all model are for single instance and do not consider the speedup that can be achieved based when training the whole ensemble on multiple GPUs.}
    \label{tab:time_and_corr}
\end{adjustbox}
\end{table}

\begin{table}[htbp]
\begin{adjustbox}{max width=\textwidth}
    \begin{tabular}{lrrr}\toprule
    \textbf{model name} & \textbf{OWA (average)} & \textbf{time to train (min)} & \textbf{ensemble size} \\
    N-BEATS (G) & 0.816 & 419 & 6\\
    N-BEATS (I) & 0.815 & 280 & 6\\
    N-BEATS parallel (G) & 0.820 & \textbf{99} & 6\\
    N-BEATS parallel (I) & 0.821 & \textbf{134} & 6\\
    \bottomrule
    \end{tabular}
    \caption{Performance of a small ensemble only trained on the MAPE loss for all lookback without bagging and time to train on a \emph{single} GPU.}
    \label{tab:smaler_ensemble}
\end{adjustbox}
\end{table}

Given the increasing trend of top-performing models requiring ever more training time \cite{makridakis2020m4}, training and deploying state-of-the-art models in real-case scenarios can entail high costs for organizations — costs that are avoidable. For instance, on Google’s cloud platform, the estimated cost of training N-BEATS(P) would drop to 530.11 USD\$ instead of the  860.13 USD\$ their price simulator gives for N-BEATS\footnote{Prices are at the rate calculated using their cost estimator on 04-08-2021, employing their "AI Platform" configuration with a single NVIDA P100 GPU}. Thus, in terms of both cost and time saved, our work provides encouraging results that suggest how multiple TS ensemble models can be accelerated without any great drawback by sharing a subset of their parameterization.

\begin{table}[htbp]
\begin{adjustbox}{max width=\textwidth}
    \begin{tabular}{lr|lrr}\toprule
        \multicolumn{1}{c}{\textbf{Model name}}&\multicolumn{1}{c}{\textbf{\# of parameters}}&\multicolumn{1}{c}{\textbf{Model name}}&\multicolumn{1}{c}{\textbf{\# of parameters}}
        \\\midrule
        N-BEATS(G) & 28'508'265'900 & N-BEATS(P, G) & \textbf{5'972'957'400} \\ 
        N-BEATS(I) & 42'288'737'310 & N-BEATS(P, I) & \textbf{8'102'076'930}\\
        N-BEATS(I+G) & 70'797'003'210 & N-BEATS(P, I+G) & \textbf{14'075'034'330} \\
        \bottomrule
    \end{tabular}
    \caption{Number of parameters for the whole ensemble for N-BEATS and N-BEATS (parallel) trained on the M4 dataset with 6 lookback windows.}
    \label{tab:params}
\end{adjustbox}
\end{table}

\section{Conclusion}
\label{sec:Conclusion}

We proposed an efficient novel architecture for training multiple TS models conjointly for univariate TS forecasting. We empirically validated the flexibility of our approach on the M4 TS datasets as well as assessing its generalizability to other domains of application, using 5 other datasets which, combined, cover over 2.5 million forecasts. We provided forecasts in various TS settings at the same level of accuracy as current state-of-the-art models with a model that is twice as fast while requiring 5 times fewer parameters than the top performing model. We highlighted both stylized facts and limitations of the performance of the model studied, in an effort to provide insights to TS practitioners for operating DNN-based models at scale. Our results suggest that training global univariate models conjointly by sharing parts of their parameterizations yield competitive forecasts in a fraction of the time and does not significantly impair either forecast accuracy or ensemble diversity.

\bibliographystyle{elsarticle-num}
\bibliography{sample.bib}


\appendix
\section{Dataset}
\label{sec:Dataset}

\begin{figure}[htbp]
    \centering
    \includegraphics[width=\linewidth]{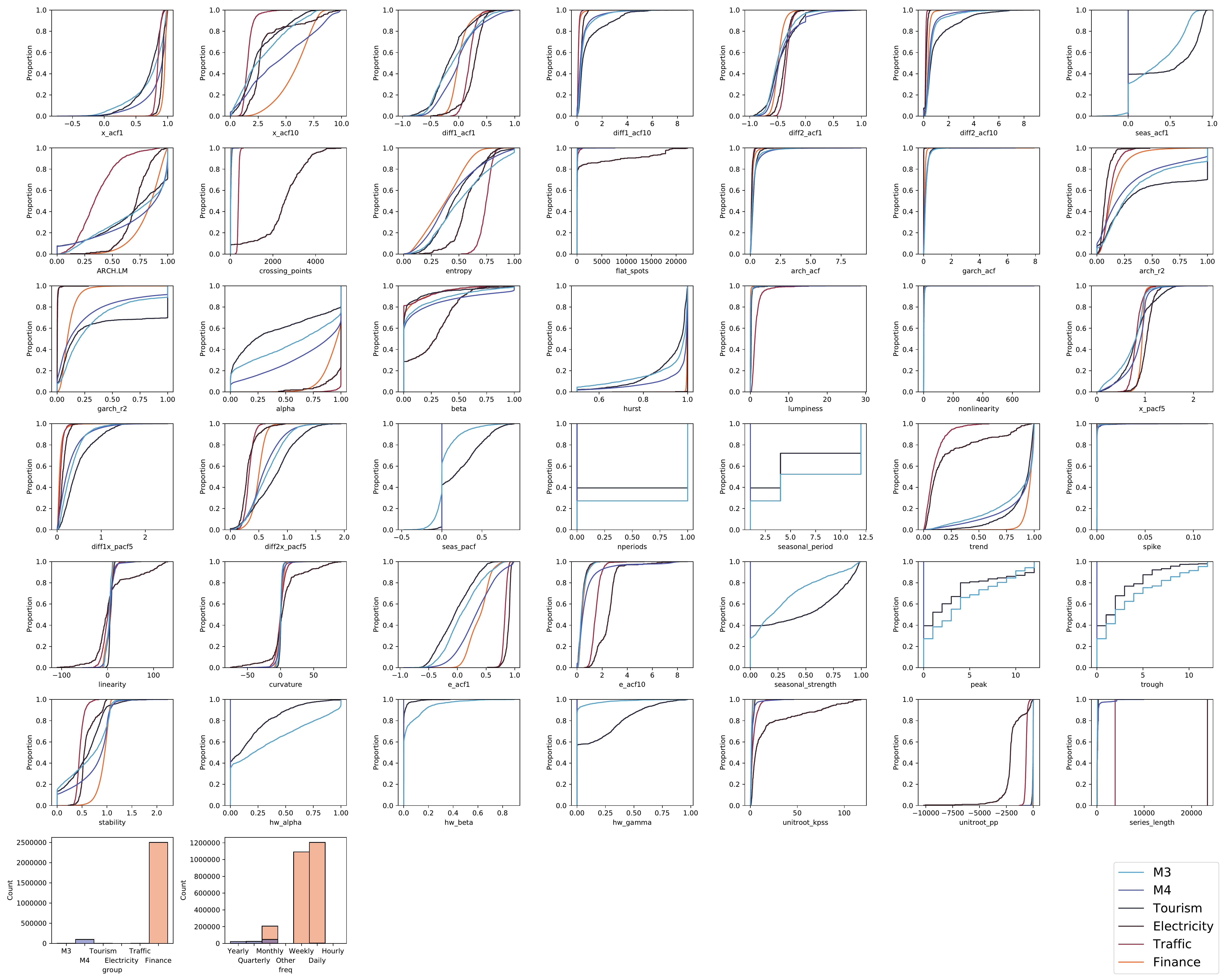}
    \caption{Cumulative distribution function plot for TS datasets over 42 statistical TS features and TS count by dataset and frequency.}
    \label{fig:cdf_plot}
\end{figure}

Fig.~\ref{fig:cdf_plot} illustrates the difference between the statistical properties of all 6 datasets, employing the same set of TS features used in the FFORMA model \cite{montero2020fforma}. We refer the reader to table.\ref{tab:features_detail} and \cite{montero2020fforma} for a detailed overview of the 42 features used and their interpretation. As an example of the observations that can be drawn from this figure: it can bee seen that both the \textbf{Electricy} and \textbf{Traffic} datasets exhibit multiple seasonal patterns, whereas datasets like \textbf{Finance} exhibit large difference from other datasets in terms of high order autocorrelation (x\_acf10), autoregressive conditional heteroscedasticity (archlm, garch\_r2), strength of trend (trend) and high variance of the mean of observation from non-overlapping windows (stability).

\begin{table}[htbp]
\begin{adjustbox}{max width=\textwidth}
    \centering
    \begin{tabular}{llp{12cm}cc}\toprule
    \multicolumn{2}{c}{\textbf{Features}}&Description&Seasonal&Non-Seasonsal \\\midrule
    1 & $T$ & length of time series & \checkmark & \checkmark\\
    2 & trend & strength of trend  & \checkmark & \checkmark\\
    3 & seasonality & strength of seasonality  & - & \checkmark\\
    4 & linearity & Linearity & \checkmark & \checkmark\\
    5 & curvature & Curvature & \checkmark & \checkmark\\
    6 & spikiness & Variance of the leave-one-out variances of the remainder component in STL decomposition & \checkmark & \checkmark\\
    7 & e_acf1 & first autocorrelation function (ACF) value of remainder series & \checkmark & \checkmark\\
    8 & e_acf10 & sum of squares of first 10 ACF values of remainder series & \checkmark & \checkmark\\
    9 & stability & sum of squares of first 10 ACF values of remainder series & \checkmark & \checkmark\\
    10 & lumpiness & Variance of the means produced for tiled (non-overlapping) windows & \checkmark & \checkmark\\
    11 & entropy & Spectral entropy (Shannon entropy) of the TS & \checkmark & \checkmark\\
    12 & hurst & Hurst exponent from \cite{engle1982autoregressive} & \checkmark & \checkmark\\
    13 & nonlinearity & Teraesvirta modified test \cite{terasvirta1994specification} & \checkmark & \checkmark \\
    13 & alpha & $\text{ETS(A,A,N)}\hat{\alpha}$ & \checkmark & \checkmark \\
    14 & beta & $\text{ETS(A,A,N)}\hat{\beta}$ & \checkmark & \checkmark \\
    15 & hwalpha & $\text{ETS(A,A,A)}\hat{\alpha}$ & \checkmark & \checkmark \\
    16 & hwbeta & $\text{ETS(A,A,A)}\hat{\beta}$ & - & \checkmark \\
    17 & hwgamma & $\text{ETS(A,A,A)}\hat{\gamma}$ & - & \checkmark \\
    18 & ur_pp & Test statistic based on Phillips-Perron test \cite{phillips1988testing} & \checkmark & \checkmark \\
    19 & ur_kpss & test statistic based on KPSS test \cite{kwiatkowski1992testing} & \checkmark & \checkmark \\
    20 & y_acf1 & first ACF value of the original series & \checkmark & \checkmark \\
    21 & diff1y_acf1 & First ACF value of the differenced series & \checkmark & \checkmark \\
    22 & diff2y_acf1 & First ACF value of the twice-differenced series & \checkmark & \checkmark \\
    23 & y_acf10 & Sum of squares of first 10 ACF values of original series & \checkmark & \checkmark \\
    24 & diff1y_acf10 & Sum of squares of first 10 ACF value of the differenced series & \checkmark & \checkmark \\
    25 & diff2y_acf10 & Sum of squares of first 10 ACF value of the twice-differenced series & \checkmark & \checkmark \\
    26 & seas_acf1 & autocorrelation coefficient at first seasonal lag & - & \checkmark \\
    27 & sediff_acf1 & first ACF value of seasonally differenced series & - & \checkmark \\
    28 & y_pacf5 & sum of squares of first 5 PACF values of original series & \checkmark & \checkmark \\
    29 & diff1y_pacf5 & sum of squares of first 5 PACF values of original series & \checkmark & \checkmark \\
    30 & diff2y_pacf5 & sum of squares of first 5 PACF values of twice-differenced series & \checkmark & \checkmark \\
    31 & seas_pacf & partial autocorrelation coefficient at first seasonal lag & \checkmark & \checkmark \\ 
    32 & crossing_points & number of times the time series crosses the median & \checkmark & \checkmark \\ 
    33 & flat_spots & number of flat spots, calculated by discretizing the series into 10 equal-sized intervals and counting the maximum run length within any single interval & \checkmark & \checkmark \\
    34 & nperiods & number of seasonal periods in the series & - & \checkmark \\
    35 & seasonal_period & length of seasonal period & - & \checkmark \\
    36 & peak & strength of peak & \checkmark & \checkmark \\
    37 & trough & strength of trough& \checkmark & \checkmark \\
    38 & ARCH.LM & ARCH.LM statistic & \checkmark & \checkmark \\
    39 & arch_acf & sum of squares of the first 12 autocorrelations of $z^2$ & \checkmark & \checkmark \\
    40 & garch_acf & sum of squares of the first 12 autocorrelations of $r^2$ & \checkmark & \checkmark \\
    41 & arch_r2 & $R^2$ value of an AR model applied to $z^2$ & \checkmark & \checkmark \\
    42 & garch_r2 & $R^2$ value of an AR model applied to $r^2$ & \checkmark & \checkmark \\
    \bottomrule
    \end{tabular}
\end{adjustbox}
\caption{List of features used to compare datasets. The functions for calculating these features are implemented in the \texttt{tsfeatures R} package by \cite{hyndman2019tsfeatures}. Default values when test failed was 0.}
\label{tab:features_detail}
\end{table}

\begin{figure}[htbp]
\centering 
\includegraphics[trim={0.1cm 0.1cm 0.1cm 0.1cm},clip,width= 0.8\textwidth]{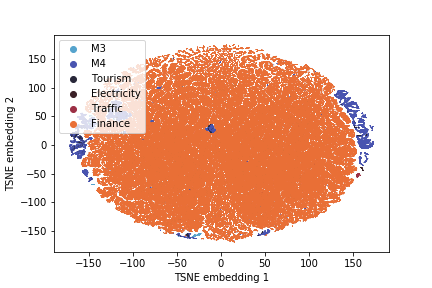}
\caption{TSNE embedding of all TS forecasted with their different subpopulations.}
\label{fig:fig1}
\end{figure}
All sampled TS from all datasets are summarized in Fig.~\ref{fig:fig1} using the T-SNE algorithm \cite{maaten2008visualizing}. Each point of this graph correspond to the 2-dimensional embedded space of a single TS computed from the same set of endogenous statistical features\cite{montero2020fforma} . One can observe the heterogeneity of these datasets and note that the subpopulations of TS within a dataset can have high variance in their statistical properties while being similar to other subpopulations of other datasets. When considering the \textbf{Finance} dataset, we can see how the behavior of the 321 TS changes is heterogenou over time in comparison to the other dataset. The \textbf{Electricity} \& \textbf{Traffic} TS share almost the same statistical properties as both populations of TS are concentrated in the same region of the graph.

\subsection{M4 Dataset Details}
\label{sec:M4 Dataset Details}

\begin{table}[htbp]
\begin{adjustbox}{max width=\textwidth}
    \centering
    \begin{tabular}{lcccccccc}\toprule
    &\multicolumn{6}{c}{\textbf{Frequency/Horizon}}&\\\cmidrule(r){2-7}
    Type & Yearly $(h=6)$ & Quarterly $(h=8)$ & Monthly $(h=18)$ & Weekly $(h=13)$ & Daily $(h=14)$ & Hourly $(h=48)$ & Total \\\midrule
    Demographic & 1'088& 1'858 & 5'728 & 24 & 10 & 0 & 8'708\\
    Finance & 6'519 & 5'305 & 10'987 & 164 & 1'559 & 0 & 24'534\\
    Industry & 3'716 & 4'673 & 10'987 & 164 & 422 & 0 & 18'798 \\
    Macro & 3'903 & 5'315 & 10'016 & 41 & 127 & 0 & 19'402 \\
    Micro & 6'538 & 6'020 & 10'975 & 112 & 1'476 & 0 & 25,121 \\
    Other & 1'236 & 865 & 277 & 12 & 633 & 414 & 3437\\\midrule
    Total & 23'000& 24'000& 48'000 & 359 & 4277 & 414 & 100'000\\
    \bottomrule
    \end{tabular}
\end{adjustbox}
\caption{Composition of the \textbf{M4} TS dataset: number of time series based on their sampling frequency and type.}
\label{tab:m4_detail}
\end{table}

The \textbf{M4}\footnote{\url{https://github.com/Mcompetitions/M4-methods}} dataset is a publicly accessible dataset that contains a large set of 100'000 heterogenous TS sampled from the \textit{ForeDeCk} database for the M4 competition\cite{makridakis2018m4}. The database is compiled at the National Technical University of Athens and is built from multiple diverse and publicly accessible sources. It includes TS frequently encountered in business domains such as industries, services, tourism, imports/exports, demographics, education, labor \& wage, government, households, bonds, stocks, insurances, loans, real estate, transportation, and natural resources \& environment. TS were sampled at different frequencies [Yearly, Quarterly, Monthly, Weekly, Daily and Hourly] each with different forecast horizons, i,e, [6, 8, 18, 13, 14, 48] according to the competition organizer. Table~\ref{tab:m4_detail} outlines the composition of the M4 dataset across domains and forecast horizons. 

All TS were provided with a prepossessing scaling procedure to ensure positive observed values at all time-steps with minimum observed values greater than or equal to 10. The scaling was applied only to sampled TS whose minimum oberved value was smaller than 10 by adding a per-TS constant to all TS to ensure that the minimal values was positive. All other TS were unaltered by any preprocessing step. The dataset was subdivized into a training and a test dataset by the M4 TS competition organizers. For further details on this dataset, we refer the reader to the following: \cite{makridakis2020m4, makridakis2018m4}. We relied on the pre-computed forecasts and PI available at \url{https://github.com/Mcompetitions/M4-methods}.

\subsection{M3 Dataset Details}
\label{sec:M3 Dataset Details}

\begin{table}[htbp]
\begin{adjustbox}{max width=\textwidth}
    \centering
    \begin{tabular}{lcccccc}\toprule
    &\multicolumn{4}{c}{\textbf{Frequency/Horizon}}&\\\cmidrule(r){2-5}
    Type & Yearly $(h=6)$ & Quarterly $(h=8)$ & Monthly $(h=18)$ & Other $(h=8)$ & Total \\\midrule
    Demographic & 245 & 57 & 111 & 0 & 413\\
    Finance & 58 & 76 & 145 & 29 & 308\\
    Industry & 102 & 83 & 334 & 0 & 519 \\
    Macro & 83 & 336 & 312 & 0 & 731 \\
    Micro & 146 & 204 & 474 & 4 & 828 \\
    Other & 11 & 0 & 52 & 141 & 204\\\midrule
    Total & 645 & 756 & 1428 & 174 & 3'003\\
    \bottomrule
    \end{tabular}
\end{adjustbox}
\caption{Composition of the \textbf{M3} TS dataset: the number of TS based on sampling frequency and type.}
\label{tab:m3_detail}
\end{table}

The \textbf{M3}\footnote{\url{https://forecasters.org/resources/time-series-data/m3-competition/}} dataset is a publicly accessible dataset that is smaller than the M4 dataset but remains relatively large and diverse. Similarly to the M4 dataset, it contains TS frequently encountered in business, financial and economic forecasting. It include yearly, quarterly, monthly, weekly, daily and hourly time series, each with different forecast horizons, i,e, [6, 8, 18, 13, 14, 48]. All series have positive observed values at all time-steps. The dataset was subdivided into a training and a test dataset by the M3 TS competition organizers. Table~\ref{tab:m3_detail} outlines the composition of the M3 dataset across domains and forecast horizons. For further details on this dataset, we refer the reader to \cite{koning2005m3}. This dataset was considered for zero-shot forecasting, to examine a case where the target dataset is from the same domains of application but with other TS.

\subsection{Tourism Dataset Details}
\label{sec:Tourism Dataset Details}

\begin{table}[htbp]
\begin{adjustbox}{max width=\textwidth}
    \centering
    \begin{tabular}{lcccc}\toprule
    &\multicolumn{3}{c}{\textbf{Frequency/Horizon}}&\\\cmidrule(r){2-4} \\
    Type & Yearly $(h=4)$ & Quarterly $(h=8)$ & Monthly $(h=24)$ & Total \\\midrule
    Tourism & 518 & 427 & 366 & 1311\\
    \bottomrule
    \end{tabular}
\end{adjustbox}
\caption{Composition of the \textbf{Tourism} TS dataset: number of time series based on sampling frequency and type.}
\label{tab:tourism_detail}
\end{table}

The \textbf{Tourism}\footnote{\url{https://robjhyndman.com/data/27-3-Athanasopoulos1.zip}} dataset is a publicly accessible dataset that contains TS collected by \cite{athanasopoulos2011tourism} from tourism government agencies and academics who had used them in previous tourism forecasting studies. The TS of this dataset are highly variable in length. It includes yearly, quarterly and monthly TS. Table.~\ref{tab:tourism_detail} details the proportion of TS from each frequency. For further detail on this dataset, we refer the reader to \cite{athanasopoulos2011tourism}. This dataset was considered for zero-shot forecasting, to examine a case where the target dataset comes from domains that are not present in the M4 dataset.

\subsection{Electricity and Traffic Datasets Details}
\label{sec:Electricity and Traffic Datasets Details}

\textbf{Electricity}\footnote{\url{https://archive.ics.uci.edu/ml/datasets/PEMS-SF}} and \textbf{Traffic} \cite{asuncion2007uci} are two publicly available datasets from the Univeristy of California Irvine Machine Learning repository. The \textbf{Electricity} dataset contains the hourly electricity usage monitoring of 370 customers over three years, with some clients being added during the the observation periods creating cold-start conditions for producing some forecasts.
The \textbf{Traffic} dataset contains TS of the hourly occupancy rates, scaled in the (0,1) range for 963 lanes of freeways in the San Francisco Bay area over a period of slightly more than a year. Both of these dataset exhibit strong seasonal patterns due to their nature and are mostly homogeneous. These two TS datasets are used de facto to evaluate the quality of DNN-based TS models as in \cite{salinas2020deepar,rangapuram2018deep,vincent2019shape}. We included these two datasets as a sanity check for zero-shot forecasting, to ensure that zero-shot forecasts were accurate in a setting where it is relatively easy to produce accurate forecasts.

\subsection{Finance dataset}
\label{sec:Finance dataset}
The \texttt{Finance} dataset contains daily closing prices of U.S. MFs and ETFs observed between 2005-07-01 and 2020-10-16 and traded on U.S. financial markets, each covering different types of asset classes including stocks, bonds, commodities, currencies and market indexes, or a proxy for a market index. The dataset was obtained through three data providers:
\begin{enumerate*}[label={(\textbf{\arabic*}})]
    \item \texttt{Fasttrack}\footnote{\url{https://investorsFasttrack.com}}, a professional-grade data provider for financial TS,
    \item \texttt{Yahoo} Finance API and
    \item the Federal Reserve of Saint-Louis (FRED) database.
\end{enumerate*}
Part of this dataset is proprietary, so we do not have permission to share that part publicly. However, the list of securities is given in Table.~\ref{tab:appendix_list_of_securities} to help interested readers reconstruct the dataset from public data sources.

We considered this dataset in our zero-shot experiments by sampling the TS at three different frequencies [dDaily, weekly and monthly] and specifying the same forecast horizon as that of the \textbf{M4} dataset. We used this dataset to present a worst-case scenario for zero-shot. First this is a case where the forecasting application is notorious for its forecasting difficulty. Moreover, the source dataset on which we train our model has at most 10K TS to train from and at worst 164 TS, which force zero-shot generalization  with very few training data. Also, by sampling the TS at large scale, we emulated how zero-shot could be applied on the whole history of the TS, similar to the procedure carried out by portfolio managers and quantitative analyst to backtest the validity of their investment strategies. The TS were split into chunks of the maximum lookback period of the N-BEATS model as training sample, and $H$ steps-ahead as testing sample.

\section{Training setup details}
\label{sec:Training setup details}

\begin{table}[htbp]
\begin{adjustbox}{max width=\textwidth}
    \centering
    \begin{tabular}{lcccccc}\toprule
    &\multicolumn{6}{c}{\textbf{Frequency/Horizon}}\\\midrule
    Frequency & Yearly $(h=6)$ & Quarterly $(h=8)$ & Monthly $(h=18)$ & Weekly $(h=13)$ & Daily $(h=14)$ & Hourly $(h=48)$ \\
    $L_H$ & 1.5 & 1.5 & 1.5 & 10 & 10 & 10\\
    Iterations N-BEATS P & 10k & 15k & 15k & 5k & 5k & 5k\\
    Iterations N-BEATS & 15k & 15k & 15k & 5k & 5k & 5k\\\cmidrule(r){2-7}
    Learning rate & \multicolumn{6}{c}{0.001}\\
    Losses & \multicolumn{6}{c}{SMAPE,MASE,MAPE}\\
    Lookback periods & \multicolumn{6}{c}{2H,3H,4H,5H,6H,7H}\\
    Batch size & \multicolumn{6}{c}{1024}\\
    Kernel size & \multicolumn{6}{c}{1}\\\cmidrule(r){2-7}
    &\multicolumn{3}{c}{\textbf{N-BEATS(G)}}&\multicolumn{3}{c}{\textbf{N-BEATS(P+G)}}\\\cmidrule(r){2-7}
    Width & \multicolumn{6}{c}{512}\\
    Blocks & \multicolumn{6}{c}{1}\\
    Blocks-layer & \multicolumn{6}{c}{4}\\
    \# Stacks &\multicolumn{6}{c}{30=[\textbf{Generic, $\cdots$, Generic}]}\\\cmidrule(r){2-7}
    &\multicolumn{3}{c}{\textbf{N-BEATS(I)}}&\multicolumn{3}{c}{\textbf{N-BEATS(P+I)}}\\\cmidrule(r){2-7}
    \# Stacks & \multicolumn{6}{c}{2 = [\textbf{Trend, Seasonality}]}\\
    T-width & \multicolumn{6}{c}{256}\\
    T-blocks & \multicolumn{6}{c}{3}\\
    T-blocks-layer & \multicolumn{6}{c}{4}\\
    S-width & \multicolumn{6}{c}{2048}\\
    S-blocks & \multicolumn{6}{c}{3}\\
    S-blocks-layer & \multicolumn{6}{c}{4}\\
    Sharing & \multicolumn{6}{c}{Stack level}\\
    \bottomrule
    \end{tabular}
\end{adjustbox}
\caption{Hyper parameters used to produce results on the \textbf{M4} TS dataset}
\label{tab:m4_hpCNN}
\end{table}
\begin{table}[htbp]
\begin{adjustbox}{max width=\textwidth}
    \centering
    \begin{tabular}{lcccccc}\toprule
    &\multicolumn{6}{c}{\textbf{Native zero-shot ($R_O$)}}\\\midrule
    Frequency & Yearly & Quarterly & Monthly & Weekly & Daily & Hourly \\
    horizon & $(h=6)$ & $(h=8)$ & $(h=18)$ & $(h=13)$ & $(h=14)$ & $(h=48)$ \\
    $L_H$ & 1.5 & 1.5 & 1.5 & 10 & 10 & 10\\
    Iterations N-BEATS & 15k & 15k & 15k & 5k & 5k & 5k\\
    Iterations N-BEATS (P) & 10k & 15k & 15k & 5k & 5k & 5k\\\midrule
    &\multicolumn{6}{c}{\textbf{Native Zero-Shot with equal forecast horizon}  ($R_{SH}$)}\\\midrule
    horizon & $(h=h^{(\mathscr{D}_{tgrt.})}_{\text{Yearly}})$ & $(h=h^{(\mathscr{D}_{tgrt.})}_{\text{Quarterly}})$ & $(h=h^{(\mathscr{D}_{tgrt.})}_{\text{Monthly}})$ & $(h=h^{(\mathscr{D}_{tgrt.})}_{\text{Weekly}})$ & $(h=h^{(\mathscr{D}_{tgrt.})}_{\text{Daily}})$ & $(h=h^{(\mathscr{D}_{tgrt.})}_{\text{Hourly}})$ \\
    $L_H$ & 1.5 & 1.5 & 1.5 & 10 & 10 & 10\\
    Iterations & 15k & 15k & 15k & 5k & 5k & 5k\\\midrule
    &\multicolumn{6}{c}{\textbf{Native Zero-Shot with equal forecast horizon}($R_{SH,LT}$)}\\\midrule
    horizon & $(h=h^{(\mathscr{D}_{tgrt.})}_{\text{Yearly}})$ & $(h=h^{(\mathscr{D}_{tgrt.})}_{\text{Quarterly}})$ & $(h=h^{(\mathscr{D}_{tgrt.})}_{\text{Monthly}})$ & $(h=h^{(\mathscr{D}_{tgrt.})}_{\text{Weekly}})$ & $(h=h^{(\mathscr{D}_{tgrt.})}_{\text{Daily}})$ & $(h=h^{(\mathscr{D}_{tgrt.})}_{\text{Hourly}})$ \\
    $L_H$ & 10 & 10 & 10 & 10 & 10 & 10\\
    Iterations & 15k & 15k & 15k & 15k & 15k & 15k\\
    \bottomrule
    \end{tabular}
\end{adjustbox}
\caption{HP differences between the different zero-shot strategies. All models were trained on the \textbf{M4} TS dataset}
\label{tab:m4_hpzero}
\end{table}

We used the same overall training framework as \cite{oreshkin2019n} including the stratified uniform sampling of TS in the source dataset to train the model. Training N-BEATS and N-BEATS(P) was done by first segmenting the training dataset into non-overlapping subsets based on the TS frequency they were observed in. Then, independent training instances were trained, one each group by specifying the forecast horizon of each instance based on the common forecast horizon of the subset. Table \ref{tab:m4_hpCNN} presents the HP settings used to train all N-BEATS and N-BEATS(P) models on the different subsets of M4. Except for the number of iterations on the yearly TS, all other HPs are the same. We did not proceed with an exhaustive parameter search since this was not the focus of our work. We were interested in whether or not we could make the N-BEATS model model faster and more usable in practical scenarios.

For zero-shot application, we relied on the scaled version of each model, i.e. where the TS is scaled based on its maximum observed value over its lookback periods. With one exception, the model trained on a given frequency split of a source dataset is used to forecast the same frequency split on the target dataset. The only exception is follows: when transferring from \textbf{M4} to \textbf{M3}, the Other subpopulation of \textbf{M3} is forecast with the model trained on the \textbf{Quarterly} subpopulation of \textbf{M4}. Table~\ref{tab:m4_hpzero} describe the different zero-shot training regimes on which the model was trained on the source dataset.

The source code to replicate the experiments for both traditional forecasting regime and zero-shot forecasting of \ref{sec:zero-shot forecasting} is available at: \url{https://anonymous.4open.science/r/actm-7F90}.

\subsection{Forecasting Combination}
\label{sec:Forecasting Combination}

Forecast combination with N-BEATS(P) and N-BEATS was done as follows: to produce a forecast from the ensemble, all forecasts of ensemble members were considered and the median was computed for every forecast for all time $t$ per TS forecast. When the forecast horizon of the model was shorter than the forecast horizon of the target dataset, we iteratively appended the forecast to the original TS signal and based our forecasts upon the transformed signal until the total forecast was longer than or equal to the forecast horizon of the target dataset. In cases where the forecast produced was longer than the forecast horizon, we truncated the forecast to keep only the $H$ first observations.

\section{Zero-shot forecasting}
\label{sec:zero-shot forecasting}

\begin{figure}[htbp]
    \centering
    \includegraphics[width=0.7\textwidth]{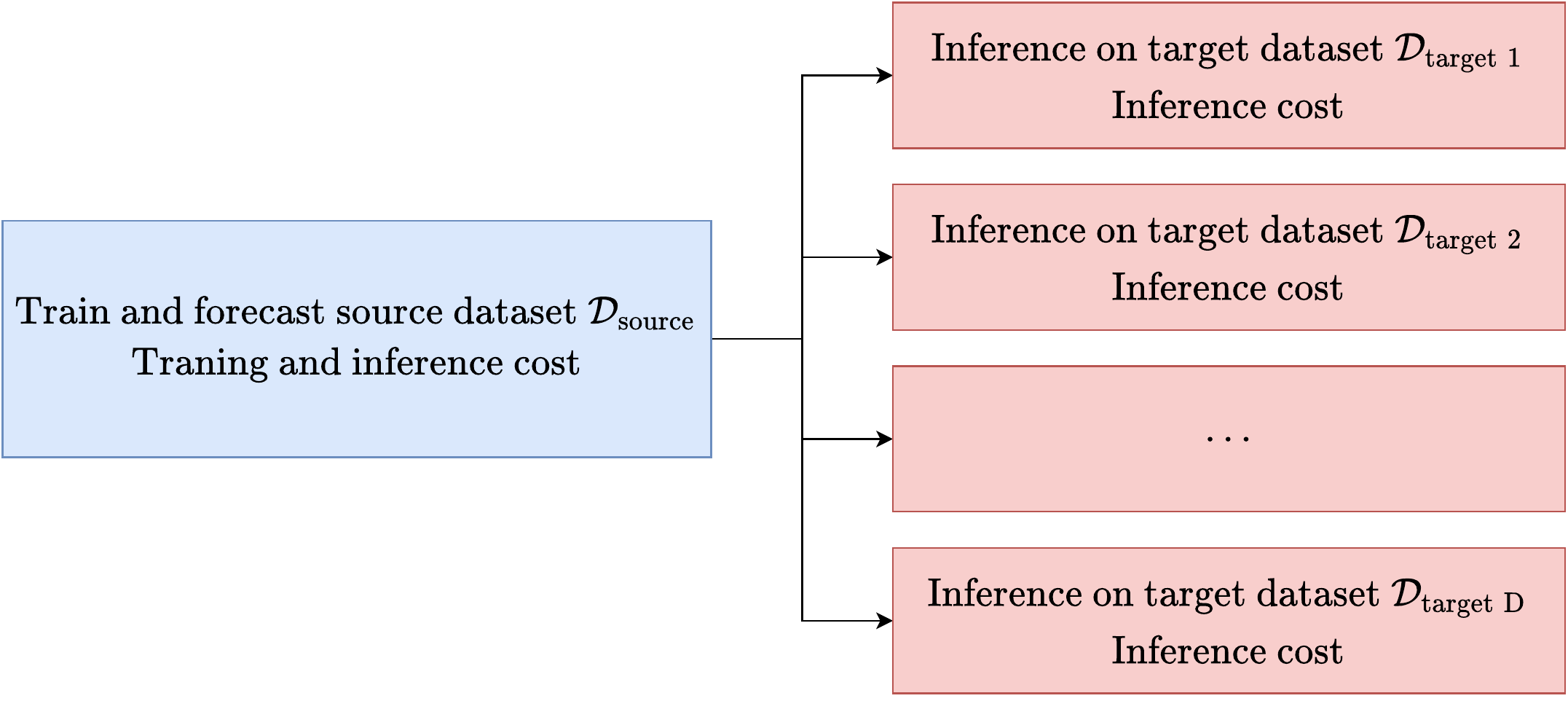}
    \caption{Flowchart of the zeros-shot forecasting regime on $D$ target datasets from one source datasets $(\mathcal{D}_{\text{source}})$. The blue square (left) represent the traditional setup of training a model and forecasting on such dataset. The red squares (red) represent the process of loading a pre-trained models and forecasting TS from a different TS dataset than the one used for training. In term of time, the time to train a model is almost always greater than the one from infering a target datasets.}
    \label{fig:zeroshotforecasting_flowchart}
\end{figure}

To test whether our model can generalize to other datasets, we evalate its capacity to support zero-shot TS forecasting, i.e., to train a neural network on a source TS dataset and deploy it on a different target TS dataset without retraining, which provides a more efficient and reliable solution to forecast at scale than its predecessor. We present a flowchart of the zero-shot forecasting regime in fig.~\ref{fig:zeroshotforecasting_flowchart}. In this setting a single model is trained once on a \textit{source} datasets and can be used to forecast multiple \textit{target} datasets without retraining as in \cite{oreshkin2020meta}. We demonstrate that N-BEATS(P) has comparable level of accuracy than N-BEATS for zero-shot generalization ability in various settings. It can operate on various domains of applications and on target datasets that are out-of-distribution of the source dataset it was trained on, i.e. on dataset from other dommains, settings and/or that have different statistical properties than the dataset it was trained on.

We evaluated their performance in the zero-shot regime on all other datasets (\textbf{M3}, \textbf{Tourism}, \textbf{Electricty}, \textbf{Traffic}, \textbf{Finance}) by training models on the \textbf{M4} dataset only using scaled TS as in \cite{oreshkin2020meta}. The reason for this preprocessing step was to prevent catastrophic failure when the target dataset scale is different from that of the source dataset.

We tested 3 different setups for zero-shot forecasting, which we denote by $R_O$, $R_{SH,LT}$ and $R_{SH}$. $R_O$ is a setup where we use the same model to produce results on $\textbf{M4}$ (Table~\ref{Tab:baseline_accuracy}) and apply it on the target dataset. This required us to truncate the forecast or apply the model iteratively on the basis of previous forecasts to ensure the forecast size is the same as the target dataset. The model was not trained to operate when this condition occurs. $R_{SH}$ is a setup where the model is trained with the same number of iterations as $R_O$ but we specified the model's forecast horizon to be the same that of the target datasets. $R_{SH,LT}$ is the same training regime as $R_{SH}$, but we allowed the model to consider TS samples from further in the past during training and trained the model with more iterations. The rationale of testing these training setup is to evaluate  the impact of training the model longer for generalization and to to test the model in forecast condition it wasn't trained to do (e.g. in $R_O$ when the forecast horizon of the target dataset exceed or is inferior to the forecast horizon of the target dataset). 

When training the model, we consider an hyper-parameter $L_H$ which is a coefficient defining the length of training history immediately preceding the last point in the train part of the TS that is used to generate training sample. This coefficient multiplied by the forecast horizon determined the maximum number of most recent points in the train dataset for each TS to generate training sample. The higher that coeffcient, the further in the past we consider TS observation to train this model. To produce forecasts, we used the subset of the ensemble models trained on the same TS frequency to produce the multiple forecasts and combined them by median aggregation. Detailed explanations of this aggregation, selection of $L_H$ and the ensemble parameters used are given in \ref{sec:Training setup details}.

We compared against statistical baselines and other ML models such as DEEP-STATE \cite{rangapuram2018deep}, N-BEATS \cite{oreshkin2019n,oreshkin2020meta}, DEEP-AR \cite{salinas2020deepar}, FFORMA \cite{montero2020fforma}, ES-RNN \cite{smyl_hybrid_2020}, Deep Factors \cite{wang2019deep} and many others including statistical baselines already evaluated on theses datasets. In reporting the accuracy of these models, we relied upon the accuracy and the pre-computed forecasts reported in their respective original paper.

\begin{table}[htbp]
\begin{adjustbox}{max width=\textwidth}
    \begin{tabular}{lllllll}\toprule
        \multicolumn{2}{c}{\textbf{M3}, SMAPE}&\multicolumn{2}{c}{\textbf{Tourism}, MAPE}&\multicolumn{2}{c}{\textbf{Electricity}, ND}&\multicolumn{1}{c}{\textbf{Traffic}, ND}\\
        \midrule
        \multicolumn{2}{l}{\textbf{N-SHOT:}} & \multicolumn{5}{c}{ }\\\cmidrule(r){1-2}
        \textit{Naive} & 16.59 & \textit{SNaive} & 24.80 & Naive & 0.37 & 0.57 \\
        \textit{Comb} \cite{makridakis2020m4} & 13.52 & \textit{ETS} \cite{hyndman2002state} & 20.88 & \textit{MatFact} \cite{yu2016temporal} & 0.16 &  0.17 \\
        \textit{ForePro} \cite{athanasopoulos2011value} & 13.19 & \textit{Theta} \cite{assimakopoulos2000theta} & 20.88 & \textit{DeepAR} \cite{salinas2020deepar} & 0.07 &  0.17 \\
        \textit{Theta} \cite{assimakopoulos2000theta} & 13.01 & \textit{ForePro} \cite{athanasopoulos2011value} & 19.84 & \textit{DeepState} \cite{rangapuram2018deep} & 0.08 &  0.17 \\
        \textit{DOTM} \cite{fiorucci2016models} & 12.90 & \textit{Strato} $\blacksquare$ & 19.52 & \textit{Theta} \cite{assimakopoulos2000theta} & 0.08 &  0.18 \\
        \textit{EXP} \cite{spiliotis2019forecasting} & 12.71 & \textit{LCBaker} \cite{baker2011winning} & 19.35 & \textit{ARIMA} \cite{box2015time} & 0.07 &  0.15 \\
        \textit{N-BEATS}\cite{oreshkin2019n} & 12.37 &  & 18.52 &  & 0.07 & 0.11 \\
        \textit{DEEP-AR*}\cite{salinas2020deepar,oreshkin2020meta} & 12.67 &  & 19.27 &  & 0.09 & 0.19 \\
        \midrule
        \multicolumn{2}{l}{\textbf{ZERO-SHOT: ($R_{SH,LT}/R_{SH}/R_O$)}} & \multicolumn{5}{c}{ }\\\cmidrule(r){1-2}
        M4 N-BEATS (G)\textit{ scaled}* \cite{oreshkin2019n} & 12.36/12.67/12.72 &  & 18.90/20.16/24.14 &  & 0.09/0.09/0.08 & 0.16/0.16/0.14 \\
        M4 N-BEATS (I)\textit{ scaled}* \cite{oreshkin2019n} & 12.43/12.63/12.66 &  & 19.43/20.58/23.26 &  & 0.10/0.09/0.08 & 0.16/0.16/0.14\\
        M4 N-BEATS (g+i)\textit{ scaled}* \cite{oreshkin2019n} & 12.38/12.61/12.64 &  & 19.04/20.22/23.43 &  & 0.10/0.09/0.08 & 0.16/0.16/0.14\\
        M4 N-BEATS (P+G)\textit{ scaled} & 12.48/12.65/12.65 &  & 18.99/19.98/22.85 &  & 0.09/0.09/0.08 & 0.16/0.18/0.14 \\
        M4 N-BEATS (P+I)\textit{ scaled} & 12.69/12.76/12.72 &  & 20.54/20.97/23.18 &  & 0.09/0.10/0.09 & 0.17/0.17/0.16 \\
        M4 N-BEATS (P+G\&I)\textit{ scaled} & 12.56/12.67/12.64 &  & 19.50/20.24/22.79 &  & 0.09/0.09/0.08 & 0.16/0.16/0.14 \\
        \bottomrule
    \end{tabular}
    \caption{Averaged forecasting for the zero-shot regime for each dataset; lower values are better. Zero-shot forecasts are compared for N-BEATS and our approach. For the models in \textit{italic} using the following references, we relied on their reported accuracy. For zero-shot results, we show the metrics for three training regimes: $R_{SH,LT}/R_{SH}/R_O$. $R_O$ is the same model used to produce the results on $\textbf{M4}$ (Table.~\ref{Tab:baseline_accuracy}), which required to truncation of the forecast or applying the model iteratively at the basis of previous forecasts to ensure the forecast size was the same that of the target dataset. $R_{SH}$ is trained in the same fashion as $R_O$ but we specified the model's forecast horizon to be the same as that of the target datasets. $R_{SH,LT}$ is the same training regime as $R_{SH}$ except that the model is allowed to consider TS samples from further in the past while  training: See Table.~\ref{tab:m4_hpzero} for more detail. Results for models with * appended to their names are replicated from the original papers and $\blacksquare$ signifies an anonymous submission for which we do not know the methodology.}
    \label{tab:zero_shot}
\end{adjustbox}
\end{table}

\begin{table}[htbp]
\begin{adjustbox}{max width=\textwidth}
    \begin{tabular}{l|ccc|ccc|ccc}\toprule
        &\multicolumn{3}{c}{\textbf{Daily} $(H=14) N=(1'222'866)$}&\multicolumn{3}{c}{\textbf{Weekly} $(H=13, N=1'091'898)$}&\multicolumn{3}{c}{\textbf{Monthly} $(H=18, N=288'114)$}\\
        Models&OWA&MDA&Time (min.)&OWA&MDA&Time (min.)&OWA&MDA&Time (min.)\\\midrule
        \textbf{N-SHOT:}\\
        Naive & 1.000 & 07.1 & --- &1.00 & 02.0 & --- & 1.000 & 00.5 & ---\\
        ARIMA \cite{box2015time} & 1.041 & 27.1 & 4685 & 1.059 & 28.9 & 3597 & \textbf{0.891} & 40.3 & 816\\
        THETA \cite{assimakopoulos2000theta} & \textbf{0.995} & 49.4 & 241 & \textbf{0.993} & 53.5 & 262 & 0.913 & \textbf{61.4} & 49\\
        SES \cite{hyndman2008forecasting} & 1.000 & 09.3 & 174 & 1.001 & 05.2 & 160 & 1.000 & 02.9 & 25\\
        HOLT \cite{chatfield1991prediction} & 1.081 & 49.6 & 167 & 1.116 & \textbf{53.9} & 160 & 0.931 & 60.3 & 42\\
        ETS \cite{hyndman2002state} & 1.019 & 20.4 & 969 & 1.044 & 18.8 & 512 & 0.940 & 29.2 & 181\\\midrule
        \textbf{ZERO-SHOT:}&\multicolumn{9}{c}{ }\\
        M4 N-BEATS (G) \cite{oreshkin2019n} & 1.165 & 50.3 & 24 & 1.078 & 50.9 & 21 & 0.963 & 55.1 & 6\\
        M4 N-BEATS (I) \cite{oreshkin2019n} & 1.222 & 50.5 & 26 & 1.045 & 51.3 & 23 & 0.962 & 54.9 & 6\\
        M4 N-BEATS (I+G) \cite{oreshkin2019n} & 1.191 & \textbf{50.7} & 50 & 1.056 & 51.1 & 44 & 0.961 & 55.4 & 12\\
        M4 N-BEATS (P+G) & 1.210 & 48.5 & 25 & 1.098 & 51.6 & 21 & 0.973 & 54.4 & 6\\
        M4 N-BEATS (P+I) & 1.135 & 48.5 & 26 & 1.055 & 52.1 & 24 & 0.975 & 54.1 & 6\\
        M4 N-BEATS (P+I\&G) & 1.139 & 49.0 & 51 & 1.044 & 52.0 & 45 & 0.973 & 54.4 & 12\\
        \bottomrule
    \end{tabular}
    \caption{Comparison between statistical baselines and zero-shot application of the N-BEATS model in terms of OWA,  MDA and time to produce forecast. Forecasts were made with the native zero-shot approach ($R_O$).}
    \label{tab:zeroshot_finance}
\end{adjustbox}
\end{table}

Table~\ref{tab:zero_shot} describes the zero-shot performance of N-BEATS and N-BEATS(P). Several observations can be made: 
\begin{enumerate}[label={\textbf{(\arabic*})}]
    \item N-BEATS(P) produces comparable zero-shot results to previous state-of-the-art models for all datasets. In other training regimes, where models trained with the same forecast horizon or longer ones, comparable levels of accuracy were observed.
    \item Comparing with \cite{oreshkin2020meta}, where a different training regime was used, the difference between their results and ours highlights the importance of the optimization procedure to facilitate transfer to another dataset. In certain cases, some datasets (e.g., \textbf{Tourism}, will benefit from a longer training, to the detriment of the forecast accuracy on the source dataset. In other cases, like the \textbf{Electricity} dataset, no adjustments are required between the source and the target dataset.
    \item The case of the \textbf{Tourism} dataset highlights the importance of ensuring that the forecast horizon of the source dataset used to train the model is longer than or equal of the target dataset; this is a key factor in producing reliable zero-shot forecasts.
\end{enumerate}
Considering that the \textbf{M4} dataset includes a large number of heterogenous TS that contain at least some TS with similar statistical properties to those present in the target dataset, zero-shot forecasting can be easily deployed with a pre-trained DNN model and can produce initial forecasts that are on par with the level of accuracy of multiple baselines, and sometimes benchmarks, very quickly at a minimal cost.

However, not all settings will benefit equally from this approach. The \textbf{Finance} dataset is a prime example of a setting where zero-shot forecasting produce mixed results. In this setting, the source TS dataset has very few TS to train from in comparison to the test sets. Also, these TS are very difficult to forecast in a univariate setting since they are almost all non-ergodic, heteroscedastic, and have high noise-to-signal ratio. Despite these added difficulties, both N-BEATS and N-BEATS(P) can produce forecasts at a comparable level to a single statistical model in term of MDA when using the $R_O$ training setup. However, the zero-shot regime achieved forecasts better than a naive one only by sampling these TS at a monthly frequency, which coincidentally is the largest pool of TS in M4 (48'000.) In comparison, the daily (3594) and weekly (227) subsets contain fewer TS. Hence, even under poor conditions of application for zero-shot N-BEATS(P), we can still produce preliminary forecasts quickly. These results highlight the importance of selecting a good source dataset but even in subpar conditions, our approach can still generalize well with respect to the MDA metric.

\section{Finance dataset: List of Securities Considered}
\label{sec:Finance dataset: List of Securities Considered}
\begin{longtable}{|p{0.3\textwidth}|p{0.4\textwidth}|p{0.31\textwidth}|}
\toprule
\multicolumn{3}{c}{\textbf{Data Source: Yahoo}}\\\midrule
\textbf{Ticker} & \textbf{Description} & \textbf{Class}\\\midrule
DJAT & Dow Jones Asian Titan 50 Index & Regional Stock Index\\
DJI & Dow Jones Industrial Average & Stock Index (US)\\
DJT & Dow Jones Transportation Average & Stock Index (US)\\
DJU & Dow Jones Utility Average & Stock Index (US)\\
GSPC & S\&P 500 & Stock Index (US)\\
IXIC & NASDAQ Composite & Stock Index (US)\\
NDX & NASDAQ-100 & Stock Index (US)\\
OEX & S\&P 100 & Stock Index (US)\\
XMI & NYSE Arca Major Market Index & Stock Index (US)\\
DX-Y.NYB & US Dollar/USDX - Index - Cash & Forex\\
FDCPX & Fidelity Select Computers & US Sector Stock Index\\
HSI & HANG SENG INDEX (Currency in HKD) & National Stock Index\\\midrule
\multicolumn{3}{c}{\textbf{Data Source: Fred}}\\\midrule
GOLDPMGBD228NLBM & Gold Fixing Price 3:00 P.M. (London time) in London Bullion Market \&  based in U.S. Dollars & Others\\
WILL4500IND & Wilshire 4500 Total Market Index & Stock Index (US)\\
WILL4500PR & Wilshire 4500 Price Index & Stock Index (US)\\
WILL5000IND & Wilshire 5000 Total Market Index & Stock Index (US)\\
WILL5000INDFC & Wilshire 5000 Total Market Full Cap Index & Stock Index (US)\\
WILL5000PR & Wilshire 5000 Price Index & Stock Index (US)\\\midrule
\multicolumn{3}{c}{\textbf{Data Source: FastTrack}}\\\midrule
FPX1 & CAC 40 Ix & National Stock Index \\
SHCP & Shanghai Composite Ix & National Stock Index \\
SPXX & STOXX Europe 600 Ix & Regional Stock Index \\
SX5P & STOXX Europe 50 Ix & Regional Stock Index \\
A-CWI & MSCI ACWI DivAdj Idx & Global Stock Index \\
A-XUS & MSCI ACWI xUS DivAdj Idx & Global Stock Index \\
AUD- & US / Australia Foreign Exchange Rate & Forex \\
BBG- & CBOE US T-Bill 13-Week Yld Bd Ix & US Bonds - Gvmnt \\
BBG-9 & BBG Barclay Agg Bond- US Universal TR Ix & US Bonds - Gvmnt \\
BBG-G & BBG Barclay Agg Bond- US Corp IG TR Ix & US Bonds - Gvmnt \\
BBG-H & ML US HY Bb-B Ix & US Bonds - Corp HY \\
BBG-I & BBG Barclay Agg Bond- US Agency Long Ix & US Bonds - Gvmnt \\
BBG-O & BBG Barclay Agg Bond- Yankee Ix & US Bonds - Gvmnt \\
BBG-S & BBG Barclay Agg Bond- US MBS Agncy TR Ix & US Bonds - Gvmnt \\
BBG-T & BBG Barclay Agg Bond- US MBS Agncy TR Ix & US Bonds - Gvmnt \\
BBG-U & BBG Muni Bond 3yr Idx & US Bonds - Gvmnt \\
BBG-Y & BBG Muni Bond 20yr Idx & US Bonds - Gvmnt \\
BBM-2 & BBG Muni Bond 5yr Idx & US Bonds - Gvmnt \\
BBM-3 & BofAML US Corp 5-7yr Total Return Ix & US Bonds - Corp Invst \\
BBM-5 & BBG Muni Bond Composite Idx & US Bonds - Gvmnt \\
BBM-I & BBG Muni Bond Long Term Idx & US Bonds - Gvmnt \\
BBM-L & BBG Muni Bond 10yr Idx & US Bonds - Gvmnt \\
BBM-T & BBG Barclay Agg Bond- US Composite TR Ix & US Bonds - Gvmnt \\
CAD- & Canada / US Foreign Exchange Rate Ix & Forex \\
CDN-X & Canadian Dollar For 100 CDN Ix & Forex \\
CHF- & Switzerland/ US Foreign Exchange Rate Ix & Forex \\
CNY- & China / US Foreign Exchange Rate Ix & Forex \\
CR-TR & CRB Total Return Ix & Commodities \\
DBC & Invesco DB Commodity Index Tracking Fund & Commodities \\
DKK- & Denmark / US Foreign Exchange Rate Ix & Forex \\
DXY-Z & US Dollar Ix & Forex \\
EFA & iShares MSCI EAFE ETF & Regional Stock Index \\
EURO- & US/Euro Foreign Exchange Rate Ix & Forex \\
EWA & iShares ETF MSCI Australia & National Stock Index \\
EWC & iShares ETF MSCI Canada & National Stock Index \\
EWD & iShares ETF MSCI Sweden & National Stock Index \\
EWG & iShares ETF MSCI Germany & National Stock Index \\
EWH & iShares ETF MSCI Hong Kong & National Stock Index \\
EWI & iShares ETF MSCI Italy Capped & National Stock Index \\
EWJ & iShares MSCI Japan ETF & National Stock Index \\
EWK & iShares ETF MSCI Belgium Capped & National Stock Index \\
EWL & iShares ETF MSCI Switzerland Capped & National Stock Index \\
EWM & iShares ETF MSCI Malaysia & National Stock Index \\
EWN & iShares ETF MSCI Netherlands & National Stock Index \\
EWO & iShares ETF MSCI Austria Capped & National Stock Index \\
EWP & iShares ETF MSCI Spain Capped & National Stock Index \\
EWS & iShares ETF MSCI Singapore & National Stock Index \\
EWW & iShares ETF MSCI Mexico Capped & National Stock Index \\
EWY-X & MSCI Korea iShr Ix & National Stock Index \\
EWZ-X & MSCI Brazil iShr Ix & National Stock Index \\
FBIOX & Fidelity Select Biotechnology & US Sector Stock Index \\
FBMPX & Fidelity Select Communication Services Portfolio & US Sector Stock Index \\
FCYIX & Fidelity Select Industrials & US Sector Stock Index \\
FDAC- & Frankfurt Dax Ix & National Stock Index \\
FDFAX & Fidelity Select Consumer Staples & US Sector Stock Index \\
FDLSX & Fidelity Select Leisure & US Sector Stock Index \\
FEZ-X & Europe 50 STOXX stTr Ix & Regional Stock Index \\
FIDSX & Fidelity Select Financial Services & US Sector Stock Index \\
FNARX & Fidelity Select Natural Resources & US Sector Stock Index \\
FNMIX & Fidelity New Markets Income  & Regional Stock Index \\
FRESX & Fidelity Fidelity Real Estate Investment Portfolio  & Others \\
FSAGX & Fidelity Select Gold & US Sector Stock Index \\
FSAIX & Fidelity Select Air Transportation & US Sector Stock Index \\
FSAVX & Fidelity Select Automotive & US Sector Stock Index \\
FSCHX & Fidelity Select Chemicals & US Sector Stock Index \\
FSCPX & Fidelity Select Consumer Discretion & US Sector Stock Index \\
FSCSX & Fidelity Select software \& Comp Service & US Sector Stock Index \\
FSDAX & Fidelity Select Defense \& Aerospace & US Sector Stock Index \\
FSDCX & Fidelity Select Commun Equipment & US Sector Stock Index \\
FSDPX & Fidelity Select Materials & US Sector Stock Index \\
FSELX & Fidelity Select Semiconductors & US Sector Stock Index \\
FSENX & Fidelity Select Energy & US Sector Stock Index \\
FSESX & Fidelity Select Energy Service & US Sector Stock Index \\
FSHCX & Fidelity Select Health Care Service & US Sector Stock Index \\
FSHOX & Fidelity Select Const \& Housing & US Sector Stock Index \\
FSLBX & Fidelity Select Brokrg \& INV Mgt & US Sector Stock Index \\
FSLEX & Fidelity Select Environmental \& Alt & US Sector Stock Index \\
FSNGX & Fidelity Select Natural Gas & US Sector Stock Index \\
FSPCX & Fidelity Select Insurance & US Sector Stock Index \\
FSPHX & Fidelity Select Health Care & US Sector Stock Index \\
FSPTX & Fidelity Select Technology & US Sector Stock Index \\
FSRBX & Fidelity Select Banking & US Sector Stock Index \\
FSRFX & Fidelity Select Transportation & US Sector Stock Index \\
FSRPX & Fidelity Select Retailing & US Sector Stock Index \\
FSTCX & Fidelity Select Telecommunications & US Sector Stock Index \\
FSUTX & Fidelity Select Utilities & US Sector Stock Index \\
FSVLX & Fidelity Select Consumer Finance & US Sector Stock Index \\
FTSE- & London FT-SE 100 Ix & National Stock Index \\
GBP- & US / UK Foreign Exchange Rate Ix & Forex \\
GLD & SPDR Gold Shares & Commodities \\
HKD- & Hong Kong / US Foreign Exchange Rate Ix & Forex \\
HY- & ML US HY Broadcastng Ix & US Bonds - Corp HY \\
HY-BC & ML US HY Build Mterl Ix & US Bonds - Corp HY \\
HY-BM & ML US HY Capitl Good Ix & US Bonds - Corp HY \\
HY-CG & ML US HY Chemicals Ix & US Bonds - Corp HY \\
HY-CH & ML US HY CCC \& Lower Ix & US Bonds - Corp HY \\
HY-CL & ML US HY Containers Ix & US Bonds - Corp HY \\
HY-CN & ML US HY Consum Prod Ix & US Bonds - Corp HY \\
HY-CP & ML US HY Div Fin Svc Ix & US Bonds - Corp HY \\
HY-DF & ML US HY Div Media Ix & US Bonds - Corp HY \\
HY-DM & ML US HY Entert Film Ix & US Bonds - Corp HY \\
HY-EF & ML US HY Energy Ix & US Bonds - Corp HY \\
HY-EG & ML US HY Environmntl Ix & US Bonds - Corp HY \\
HY-EN & ML US HY Ex Telecom Ix & US Bonds - Corp HY \\
HY-ET & ML US HY Fd Bvrge Tb Ix & US Bonds - Corp HY \\
HY-FB & ML US HY Fd\&Drg Retl Ix & US Bonds - Corp HY \\
HY-FD & ML US HY Homebldr Re Ix & US Bonds - Corp HY \\
HY-HB & ML US HY Healthcare Ix & US Bonds - Corp HY \\
HY-HC & ML US HY Insurance Ix & US Bonds - Corp HY \\
HY-IN & ML US HY Leisure Ix & US Bonds - Corp HY \\
HY-LE & ML US HY Metal Minng Ix & US Bonds - Corp HY \\
HY-MM & ML US HY Paper Ix & US Bonds - Corp HY \\
HY-PP & ML US HY Publsh Prnt Ix & US Bonds - Corp HY \\
HY-PR & ML US HY Restaurants Ix & US Bonds - Corp HY \\
HY-RS & ML US HY Super Retl Ix & US Bonds - Corp HY \\
HY-SR & ML US HY Steel Ix & US Bonds - Corp HY \\
HY-ST & ML US HY Services Ix & US Bonds - Corp HY \\
HY-SV & ML US HY Tech\&Aerosp Ix & US Bonds - Corp HY \\
HY-TA & ML US HY Telecommnct Ix & US Bonds - Corp HY \\
HY-TC & ML US HY Cabl Sat Tv Ix & US Bonds - Corp HY \\
HY-TV & ML US HY Utilities Ix & US Bonds - Corp HY \\
HY-UT & ML US Indl Corps A Ix & US Bonds - Corp Invst \\
IC-1A & ML US Indl Corps AA Ix & US Bonds - Corp Invst \\
IC-2A & ML US Indl Corps AAA Ix & US Bonds - Corp Invst \\
IC-3A & ML US Indl Corps BBB Ix & US Bonds - Corp Invst \\
IEF & iShares ETF 7 10 Year Treasury Bond & US Bonds - Gvmnt \\
INE-X & MSCI Italy iShr Ix & National Stock Index \\
INH-X & MSCI Hong Kong iShr Ix & National Stock Index \\
INR- & India/ US Foreign Exchange Rate Ix & Forex \\
INR-X & MSCI Singapore iShr Ix & National Stock Index \\
IWC-X & Russell Microcap & Stock Index (US) \\
IXF-X & NASDAQ Financial-100 & Stock Index (US) \\
JPY- & Japan/ US Foreign Exchange Rate Ix & Forex \\
KRW- & South Korea / US Exchange Rate Ix & Forex \\
LLQ-X & Russell Microcap - Dividend Adj & Stock Index (US) \\
LLR-X & Russell Microcap & Stock Index (US) \\
LQD & iShares iBoxx \$ Investment Grade Corporate Bond ETF & US Bonds - Corp Invst \\
M-BRC & MSCI Emerging Markets BRIC DivAdj Idx & Global Stock Index \\
M-CN & MSCI China DivAdj Ix & National Stock Index \\
M-DEO & MSCI Dev Mkts Euro DivAdj Idx & Global Stock Index \\
M-WD & MSCI World DivAdj Idx & Global Stock Index \\
M16Y- & BofAML US Corporate A Effective Yield Ix & US Bonds - Corp Invst \\
M26Y- & BofAML US Corporate AA Effective Yield I & US Bonds - Corp Invst \\
M36Y- & BofAML US Corporate AAA Effective Yield & US Bonds - Corp Invst \\
M3EY- & BofAML US Corp AAA Option-Adj Spread Ix & US Bonds - Corp Invst \\
M46Y- & BofAML US Corporate BBB Effective Yield & US Bonds - Corp Invst \\
M56Y- & BofAML US Corporate 1-3 Year Effective Y & US Bonds - Corp Invst \\
M5EY- & BofAML US Corp 1-3Y Option-Adj Spread Ix & US Bonds - Corp Invst \\
M66Y- & BofAML US Corporate 3-5 Year Effective Y & US Bonds - Corp Invst \\
M6EY- & BofAML US Corp 3-5Y Option-Adj Spread Ix & US Bonds - Corp Invst \\
M76Y- & BofAML US Corporate 5-7 Year Effective Y & US Bonds - Corp Invst \\
M7TR- & BBG Muni Bond 7yr Idx & US Bonds - Gvmnt \\
M86Y- & BofAML US Corporate 7-10 Year Effective & US Bonds - Corp Invst \\
M8TR- & BBG Muni Bond 1yr Idx & US Bonds - Corp Invst \\
M96Y- & BofAML US Corporate 10-15 Year Effective & US Bonds - Corp Invst \\
M9EY- & BofAML US HY BB Option-Adj Spread Ix & US Bonds - Corp HY \\
MDY & StateSt ETF SPDR S\&P MIDCAP 400 & Stock Index (US) \\
MF6Y- & BofAML US Corporate 15 Year Effective Yi & US Bonds - Corp Invst \\
MFEY- & ML US T-Bill 0-3mo Div-Adj Ix & US Bonds - Gvmnt \\
ML-03 & ML US T-Bill 1-10yrs Ix & US Bonds - Gvmnt \\
ML-10 & ML US T-Bill 1-3yrs Div-Adj Ix & US Bonds - Gvmnt \\
ML-13 & ML US T-Bill 12mo Div-Adj Ix & US Bonds - Gvmnt \\
ML-1Y & ML US T-Bill 3-5yrs Div-Adj Ix & US Bonds - Gvmnt \\
ML-35 & ML US T-Bill 3-6mo Div-Adj Ix & US Bonds - Gvmnt \\
ML-36 & ML US T-Bill 6mo Div-Adj Ix & US Bonds - Gvmnt \\
ML-6T & ML US T-Bill 7-10yrs Ix & US Bonds - Gvmnt \\
ML-70 & BofAML US High Yield B Total Return Inde & US Bonds - Corp HY \\
ML-I0 & ML US T-Bill 1-10yrs Infl-Lnk Ix & US Bonds - Gvmnt \\
ML-I1 & BBG Barclay Agg Bond- TBill Tips TR Ix & US Bonds - Gvmnt \\
ML-TB & ML US Corp Non-Fd\&Dru Ret Ix & US Bonds - Corp Invst \\
MLB- & BofAML US High Yield BB Total Return Ind & US Bonds - Corp HY \\
MLBB- & BofAML US High Yield CCC or Below Total & US Bonds - Corp HY \\
MLCC- & ML US HY Master II D-A H0A0 Ix & US Bonds - Corp HY \\
MLHY- & ML BBB Grade Div-Adj Muni Ix & US Bonds - Gvmnt \\
MLMB- & ML Municipal Master Div-Adj Ix & US Bonds - Gvmnt \\
MLMM- & ML US T-Bill Div-Adj Ix & US Bonds - Gvmnt \\
MXN- & Mexico / US Foreign Exchange Rate Ix & Forex \\
OSX-X & AMEX Oil Service HLDRS Ix & Commodities \\
PCY & Invesco Emerging Markets Sovereign Debt ETF \& US Bonds - Corp HY \\
RTF-X & Russell Top 50 & Stock Index (US) \\
RU2-D & Russell 2500 - Dividend Adj & Stock Index (US) \\
RUA-D & Russell 3000 - Dividend Adj & Stock Index (US) \\
RUA-X & Russell 3000 & Stock Index (US) \\
RUI-D & Russell 1000 - Dividend Adj & Stock Index (US) \\
RUI-I & Russell 1000 & Stock Index (US) \\
RUM-D & Russell MidCap - Dividend Adj & Stock Index (US) \\
RUP-D & Russell Top 200 - Dividend Adj & Stock Index (US) \\
RUP-X & Russell Top 200 & Stock Index (US) \\
RUS-D & Russell Small Cap - Dividend Adj & Stock Index (US) \\
RUT-D & Russell 2000 - Dividend Adj & Stock Index (US) \\
RUT-U & Russell 2000 - Unadj & Stock Index (US) \\
S-100 & S\&P Global 100 Ix & Global Stock Index \\
SEK- & Sweden / US Foreign Exchange Rate Ix & Forex \\
SGD- & Singapore / US Foreign Exchange Rate Ix & Forex \\
SHY & iShares 1-3 Year Treasury Bond ETF & US Bonds - Gvmnt \\
THB- & Thailand / US Foreign Exchange Rate Ix & Forex \\
TIP & iShares TIPS Bond ETF & US Bonds - Gvmnt \\
TLT & iShares 20+ Year Treasury Bond ETF & US Bonds - Gvmnt \\
TWD- & Taiwan / US Foreign Exchange Rate Ix & Forex \\
UC- & ML US Corp 10 Yrs Ix & US Bonds - Corp Invst \\
UC-10 & ML US Corp 15 Yrs Ix & US Bonds - Corp Invst \\
UC-15 & ML US Corp Gs\&Elct Utl 1-10 Yrs Ix & US Bonds - Corp Invst \\
UC-G1 & ML US Corp Gas\&Elect Utl Ix & US Bonds - Corp Invst \\
UC-G4 & ML US Corp Phones 10-15 Yrs Ix & US Bonds - Corp Invst \\
UC-LC & ML US T-Bill 7-10yrs Infl-Lnk Ix & US Bonds - Gvmnt \\
UC-P1 & ML US Corp Phones 15 Yrs Ix & US Bonds - Corp Invst \\
UC-P2 & ML US Corp Utils\&Phones Ix & US Bonds - Corp Invst \\
UC-UP & ML US Corp Large Cap Ix & US Bonds - Corp Invst \\
US05- & BofAML US Corporate 7-10yr Total Return & US Bonds - Corp Invst \\
UUP & Invesco DB US Dollar Index Bullish Fund & Forex \\
VASVX & Vanguard Selected Value Fund & Stock Index (US) \\
VBISX & Vanguard Short Term Bond Index & US Bonds - Gvmnt \\
VEIEX & Vanguard Emerging Market Stock Index INV & Regional Stock Index \\
VEXMX & Vanguard Extended Market Index Fund  & Global Stock Index \\
VEXPX & Vanguard Explorer Fund INV & Stock Index (US) \\
VFICX & Vanguard Int. Term Investment Grade Bond Fund & US Bonds - Corp Invst \\
VFIIX & Vanguard GNMA INV & US Bonds - Gvmnt \\
VFISX & Vanguard Short-Term Treasury INV  & US Bonds - Gvmnt \\
VFITX & Vanguard Intermediate Term Treasury Fund & US Bonds - Gvmnt \\
VFSTX & Vanguard Short-Term INV Growth Incm INV & US Bonds - Corp Invst \\
VGENX & Vanguard Energy INV & National Stock Index \\
VGHCX & Vanguard Health Care INV & National Stock Index \\
VGPMX & Vanguard Global Capital Cycles Fund & Stock Index (US) \\
VGSIX & Vanguard REIT Index INV & Others \\
VINEX & Vanguard International Explorer Fund & Global Stock Index \\
VNQ & Vanguard Real Estate Index Fund ETF Shares & Others \\
VTRIX & Vanguard International Value Fund  & Global Stock Index \\
VTSMX & Vanguard Total Stock Markets Index INV & Global Stock Index \\
VUSTX & Vanguard Long-Term Treasury INV & US Bonds - Gvmnt \\
VWEHX & Vanguard Hi-Yield Corporate INV & US Bonds - Corp HY \\
VWESX & Vanguard Long-Term INV Growth Income INV & US Bonds - Corp Invst \\
VWIGX & Vanguard International Growth INV & Others \\
VWINX & Vanguard Wellesley Income INV & US Bonds - Gvmnt \\
VWO & Vanguard FTSE Emerging Markets Index Fund ETF Shares & Global Stock Index \\
VWUSX & Vanguard US Growth INV & Stock Index (US) \\
VXF & Vanguard Extended Market Index Fund ETF Shares & Global Stock Index \\
WDG-X & MSCI Germany iShr Ix & National Stock Index \\
WPB-X & MSCI Canada iShr Ix & National Stock Index \\
XLB & StateSt ETF Materials Select Sector SPDR & US Sector Stock Index \\
XLE & StateSt ETF Energy Select Sector SPDR Fd & US Sector Stock Index \\
XLF & StateSt ETF Financial Select Sector SPDR & US Sector Stock Index \\
XLI & StateSt ETF Industrial Sel Sector SPDR & US Sector Stock Index \\
XLK & StateSt ETF Tech Select Sector SPDR & US Sector Stock Index \\
XLP & StateSt ETF Consumer Staples SelSctrSPDR & US Sector Stock Index \\
XLU & StateSt ETF Utilities Select Sector SPDR & US Sector Stock Index \\
XLV & StateSt ETF Health Care Sel Sector SPDR & US Sector Stock Index \\
XLY & StateSt ETF Consumer DiscretnrySlSctSPDR & US Sector Stock Index \\
XLC & StateSt ETF Communication Service SlSctSPDR & US Sector Stock Index \\
XLRE & StateSt ETF Real Estate  SlSctSPDR & US Sector Stock Index \\
VOX & Vanguard Communication Services Index Fund ETF Shares & US Sector Stock Index \\
IYR & iShares U.S. Real Estate ETF & US Sector Stock Index \\
XOI-I & AMEX Oil Ix & Commodities \\
ZAR- & South Africa/ US Exchange Rate Ix & Forex \\
NIKI & Tokyo Nikkei Ix & National Stock Index \\
BBM-1 & BBG Muni Bond 1-10yr Blend Idx & US Bonds - Corp Invst \\
BBM-7 & BBG Muni Bond 15yr Idx & US Bonds - Gvmnt \\
BBM-B & BBG Barclay Agg Bond- Lng Govt/Crd TR Ix & US Bonds - Corp Invst \\
BBM-F & US Treasury 5-Year Bd Yield Ix & US Bonds - Gvmnt \\
BRL- & Brazil / US Foreign Exchange Rate Ix & Forex \\
DIA & StateSt ETF SPDR Dow Jones IndustrilAvrg & Stock Index (US) \\
DJ-CO & DJ UBS Crude Oil Ix & Commodities \\
EWQ & iShares ETF MSCI France & National Stock Index \\
EWU & iShares ETF MSCI United Kingdom & National Stock Index \\
IC-3B & BofAML US Corporate A Semi-Annual Yield & US Bonds - Corp Invst \\
IEO-X & DJ US Oil \& Gas iShr Ix & Commodities \\
M-CNA & MSCI China A DivAdj Ix & National Stock Index \\
M-DEA & MSCI EAFE DivAdj Idx & Global Stock Index \\
M-DEU & MSCI Dev Mkts EU DivAdj Idx & Regional Stock Index \\
M-DG7 & MSCI Dev Mkts G7 Index DivAdj Idx & Global Stock Index \\
M-EM & MSCI Emerging Markets DivAdj Idx & Global Stock Index \\
M-EMA & MSCI Emerging Markets Asia DivAdj Idx & Regional Stock Index \\
M-EME & MSCI Emergin Martkets EMEA DivAdj Idx & Regional Stock Index \\
M-EMU & MSCI Emerging Markets Europe DivAdj Idx & Regional Stock Index \\
M1EY- & BofAML US Corporate AA Semi-Annual Yield & US Bonds - Corp Invst \\
M2EY- & BofAML US Corporate AAA Semi-Annual Yiel & US Bonds - Corp Invst \\
M3OA- & BofAML US Corporate BBB Semi-Annual Yiel & US Bonds - Corp Invst \\
M4EY- & BofAML US Corporate 1-3 Year Semi-Annual & US Bonds - Corp Invst \\
M5OA- & BofAML US Corporate 3-5 Year Semi-Annual & US Bonds - Corp Invst \\
M6OA- & BofAML US Corporate 5-7 Year Semi-Annual & US Bonds - Corp Invst \\
M7EY- & BofAML US Corporate 7-10 Year Semi-Annua & US Bonds - Corp Invst \\
M8EY- & BofAML US Corporate 10-15 Year Semi-Annu & US Bonds - Corp Invst \\
MBOA- & BofAML US Corporate 15 Year Semi-Annual & US Bonds - Corp Invst \\
QQQ & Nasdaq 100 ETF & Stock Index (US) \\
SP-GB & S\&P Global BMI Idx DivAdj & Global Stock Index \\
SP-GL & S\&P Global 1200 Idx DivAdj & Global Stock Index \\
SP-HB & S\&P 500 High Beta Idx DivAdj & Global Stock Index \\
SP-IO & S\&P Global 100 Idx DivAdj & Global Stock Index \\
SP-L4 & S\&P Latin America 40 Idx DivAdj & Regional Stock Index \\
SPY & StateSt ETF SPDR S\&P 500 & Stock Index (US) \\
ST-AG & Silver Spot & Commodities \\
ST-AU & Gold Spot & Commodities \\
ST-BC & Brent Crude Spot & Commodities \\
ST-CA & Cocoa Spot & Commodities \\
ST-CF & Coffee Bushel Spot & Commodities \\
ST-CO & Crude Oil Spot & Commodities \\
ST-CT & Cotton Bushel Spot & Commodities \\
ST-CU & Copper Spot & Commodities \\
ST-HO & Heating Oil Spot & Commodities \\
ST-NG & Natural Gas Spot & Commodities \\
ST-PD & Palladium Spot & Commodities \\
ST-PL & Platinum Spot & Commodities \\
WTI-B & Blmbrg WTI Crude Oil Sub Ix Total Return & Commodities \\
VIPSX & Vanguard Inflation-Protected Securities Fund Investor Shares & US Bonds - Gvmnt
\\\bottomrule
\caption{List of US traded funds used to create the \textbf{finance} dataset. The class columns correspond to the type of securities and the source columns specify where the TS was collected. See Tab.~\ref{tab:appendix_list_of_securities2} for a brief description of the asset classes}
\label{tab:appendix_list_of_securities}
\end{longtable}

\begin{table}[htbp]
\begin{adjustbox}{max width=\textwidth}
\begin{tabular}{c|c}\toprule
\textbf{TS type} & \textbf{Description} \\\midrule
US Stock Index	& Index of US stocks, such as the S\&P500  \\
US Stock &	A fund (ETF or mutual funds) made up primarily of US stocks \\
US Sector Stock Index &	US stock industry sector index \\
Regional Stock Index & Global region stock index, such as Europe \\
National Stock Index & Country stock index \\
Global Stock Index & Global / world stock index \\
US Bonds - Gvmnt & US treasury funds \\
US Bonds - Corp Invst & US corporate bond funds, investment grade \\
US Bonds - Corp HY & US bond funds, high yield \\
Country Funds & Country index fund \\
Forex & Foreign Exchange \\
Commodities & Commodities tracking fund \\
Real Estate & Real estate fund \\
Other & Other fund or index \\\bottomrule
\end{tabular}
\caption{Brief description of the typea of TS used in the \textbf{Finance} dataset.}
\label{tab:appendix_list_of_securities2}
\end{adjustbox}
\end{table}

\end{document}